\documentclass[10pt,twocolumn,letterpaper]{article}

\usepackage{iccv}
\usepackage{times}
\usepackage{epsfig}
\usepackage{graphicx}
\usepackage{amsmath}
\usepackage{amssymb}
\usepackage{amsfonts,amssymb} 
\usepackage{multirow}
\usepackage{subfigure}
\usepackage{caption}
\usepackage{graphicx}
\usepackage{float} 
\usepackage{pdfpages}
\usepackage{array}

\usepackage[breaklinks=true,bookmarks=false]{hyperref}
\usepackage{float}
\iccvfinalcopy % *** Uncomment this line for the final submission

\def\iccvPaperID{****} % *** Enter the ICCV Paper ID here

\ificcvfinal\fi
\begin{document}

%%%%%%%%% TITLE
\title{\ Class-level  Multiple Distributions Representation  are Necessary for Semantic Segmentation}
\author{Jianjian Yin${^{1,2}}$  \quad \quad
%\and
Zhichao Zheng${^{2,3}}$\quad \quad
%{\tt\small zheng\underline{ }zhichaoX@163.com}
%\and
Yanhui Gu${^{1,2}}$  \quad \quad
%\and
Junsheng Zhou${^{1,2}}$  \quad  \quad
%\and
Yi Chen${^{1,2}}$\thanks{Corresponding author.}\\
%\and
 $^2$School of Computer and Electronic Information, Nanjing Normal University, China\\ 
$^1$\{\tt\small{212202015}, gu,  zhoujs, cs\underline{ }chenyi\}@njnu.edu.cn  \quad
 $^3$zheng\underline{ }zhichaoX@163.com  \\
}

\maketitle
\ificcvfinal\thispagestyle{empty}\fi

%%%%%%%%% ABSTRACT
\begin{abstract}
   Existing approaches focus on using class-level features to improve semantic segmentation performance. How to characterize the relationships of intra-class pixels and inter-class pixels is the key to extract the discriminative representative class-level features.    In this paper,  we introduce for the first time   to describe  intra-class variations by multiple distributions. Then, multiple distributions representation learning(\textbf{MDRL}) is proposed  to augment the pixel representations for semantic segmentation.  Meanwhile, we design a class multiple distributions consistency strategy to construct discriminative multiple distribution representations of embedded pixels. Moreover, we put forward a multiple distribution semantic aggregation module to aggregate multiple distributions of the corresponding class to enhance pixel semantic information.   Our approach can be seamlessly integrated into popular segmentation frameworks FCN/PSPNet/CCNet and achieve  5.61\%/1.75\%/0.75\% mIoU improvements on ADE20K.   Extensive experiments on the Cityscapes, ADE20K datasets have proved that our method can bring significant performance improvement.
\end{abstract}

%%%%%%%%% BODY TEXT
\section{Introduction}

Semantic segmentation is a classical task in computer vision, aiming  to assign semantic labels to each pixel in an image accurately. Semantic segmentation has extensive applications in the fields of autonomous driving, medical diagnosis and so on. In the last few years, the performance of semantic segmentation has improved tremendously with the development of deep neural network\cite{he2016deep},\cite{chen2017deeplab}.  Full convolutional network\cite{long2015fully} based on encoder-decoder architecture becomes the cornerstone of all approaches. Existing methods generally focus on two problems to improve the performance of segmentation. One is how to change the structure of the network so as to improve the pixel feature representation\cite{wang2020deep},\cite{chen2018encoder},\cite{yu2017dilated},\cite{chen2017rethinking}. The other is to obtain contextual information to enhance the pixel representation\cite{chen2017deeplab},\cite{fu2019dual},\cite{he2019dynamic},\cite{huang2019ccnet},\cite{yuan2018ocnet},\cite{zhao2017pyramid}. This paper investigates the same direction as the latter, with the aim of how to obtain richer contextual/semantic  information to improve the performance of segmentation.

\begin{figure}[t]
   \centering

   \subfigure[FCN]{
		\begin{minipage}[t]{0.49\linewidth}
			\centering
			\includegraphics[width=0.95\linewidth,height=2.5cm]{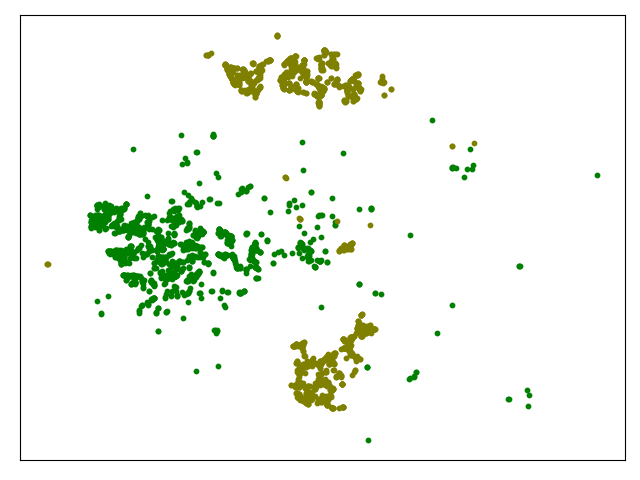}\\
		\end{minipage}%
	}%
 \subfigure[FCN+MDRL]{
		\begin{minipage}[t]{0.49\linewidth}
			\centering
			\includegraphics[width=0.95\linewidth,height=2.5cm]{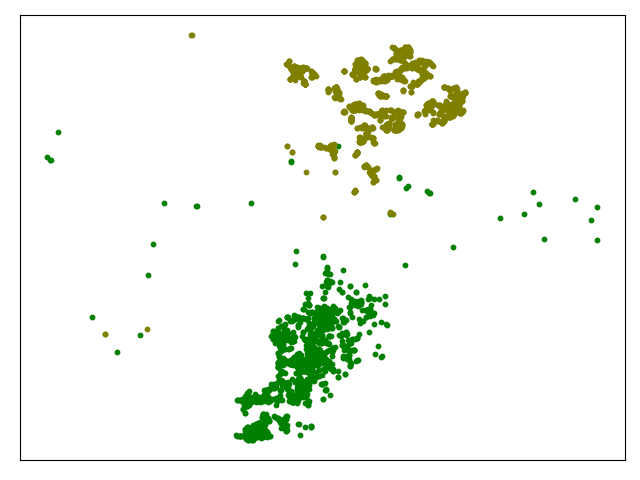}\\
		\end{minipage}%
	}% 
   \caption{ \textbf{t-SNE visualization on Cityscapes val.} We selected pixel features of the train(olive points) and road(green points)  for tsne visualization. }
   \label{fig:1}
\vspace{-1.2em}
\end{figure}

 The methods obtaining contextual information are broadly divided into two types, multi-scale feature aggregation and relational contextual aggregation. For multi-scale feature aggregation, Deeplab\cite{chen2017deeplab} uses various dilation convolutions to capture contextual information at multiple scales. PSPNet\cite{zhao2017pyramid} introduces pyramid spatial pooling to aggregate contextual information. For relational contextual aggregation,  ACFNet\cite{zhang2019acfnet} and OCRNet\cite{yuan2020object} divide pixels in an image into multiple regions and then increase the pixel representation by weighting the aggregated region representation, the weights are determined by the relationship between the pixels and  the regions. Although the above methods are effective, they ignore the potential contextual information between the input images. In other words,there is no consideration of using class-level features beyond image  to enhance the pixel representations.

     In order  to obtain class-level features  beyond the input image, MCIBI\cite{jin2021mining}
proposes to use simulated annealing to find a  semantic feature on each class. Furthermore,  MCIBI++\cite{jin2022mcibi++} assuming that the class-level features satisfy a \textbf{Gaussian} distribution, the pixel representations are enhanced by randomly generating the corresponding class feature by mean and variance each time.  As shown in the Figure \ref{fig:1},we listed pixel features of  two classes for tsne visualization, the result indicates that intra-class features hava large divergence.  MCIBI\cite{jin2021mining} and  MCIBI++ \cite{jin2022mcibi++}  both use a feature or distribution to describe the representation of each class, which is not comprehensive. 
\begin{figure}[t]
  \centering
   \includegraphics[width=\linewidth]{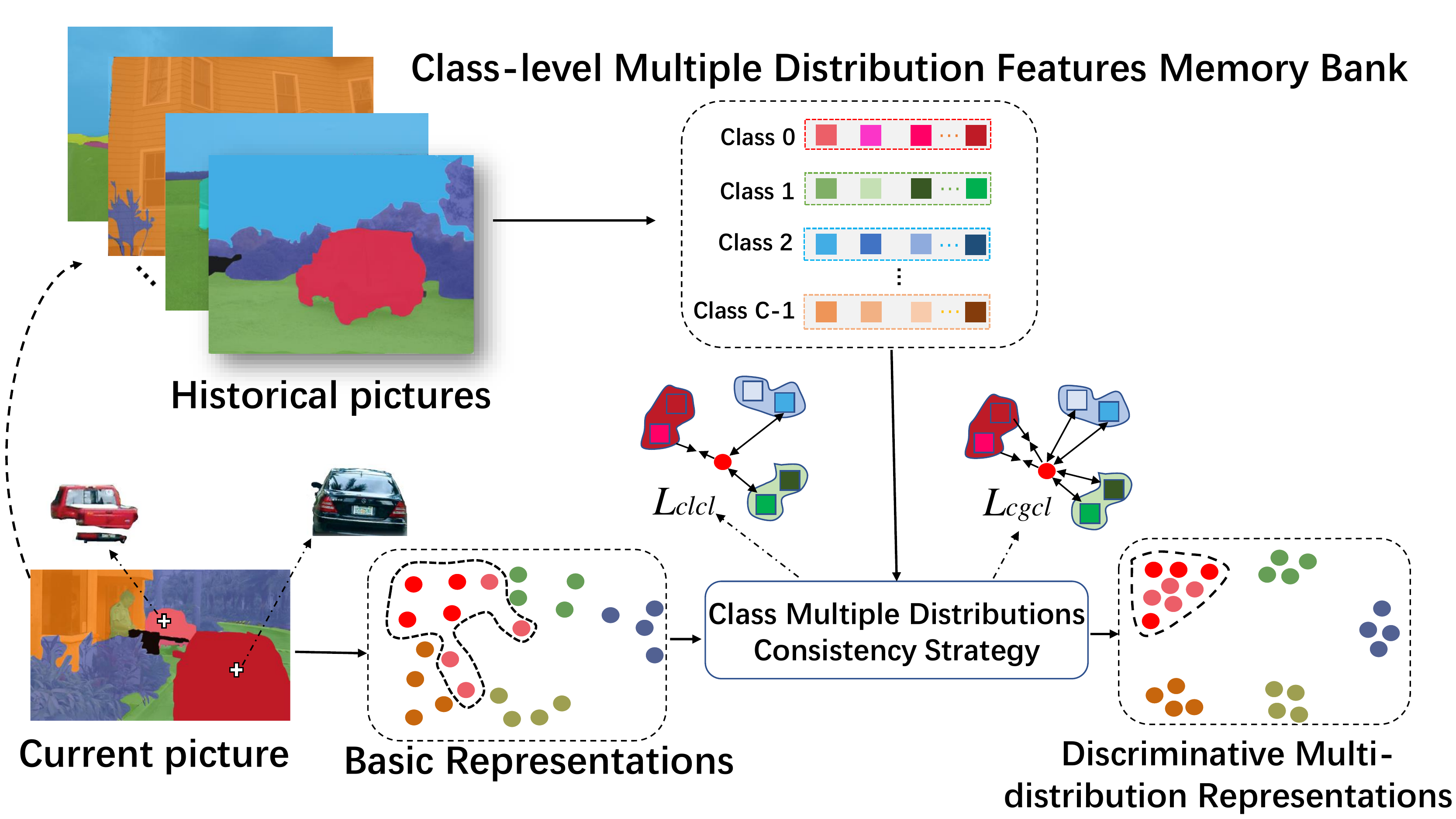}
  \caption{ \textbf{Describe  class representation by multiple distribution features.}The representation of cars with different colors are obviously different. Intra-class compact and inter-class dispersion  are realized by class multiple distributions consistency strategy, which includes the class local consistency loss ${ L_{clcl}}$ and  the class global consistency loss ${L_{cgcl}}$. } 
    \label{fig:2}

\end{figure}

In this paper,multiple distributions are introduced  to describe intra-class divergence.  We propose the multiple distributions representation learning(MDRL) to extract more semantic information per class, and augment the pixel representations.  Meanwhile, we design the class multiple distributions consistency strategy, which contains the class local/global consistency loss. The class local consistency loss is to make pixels more compact with the nearest distribution of  the same class  and father from the other classes nearset distribution.  Class global consistency loss leverages multiple distribution features per class to achieve intra-class compactness and inter-class dispersion. Details can be seen in Figure \ref{fig:2}.  Several single distribution  feature representations are obtained by feature voting based on the weights of the pixel representations  and  multiple distribution features for each class.  Finally, a novel multiple distributions semantic aggregation module aggregates several single distribution feature representations to get  the  more fine-grained multiple distribution feature representations. 

In a nutshell, our main contributions have four main points:

$\bullet$To the best of our knowledge, this paper  first  explores  multiple distributions  to describe intra-class variations, resulting in richer class-level semantic features.

$\bullet$We propose a class multiple distributions consistency strategy to construct  discriminative multiple distribution feature representations of embedded pixels. 

$\bullet$We put forward a  noval  multiple distributions semantic aggregation module to obtain more fine-grained multiple distribution feature representations, which is  used to augment the pixel representations.

$\bullet$We have done extensive experiments on different datasets and compared it with other state-of-the-art methods, the results show that our method can bring significant performance improvement.

\section{Related Work}
\textbf{Semantic Segmentation}  Semantic segmentation methods based on FCN\cite{long2015fully} have been very successful. Some studies\cite{chen2017deeplab},\cite{zheng2015conditional},\cite{liu2017deep} have targeted the refinement of the results generated by FCN. Specifically, these studies are broadly divided into two directions. One is to design more efficient network\cite{yu2017dilated},\cite{wang2020deep},\cite{zhang2022resnest},\cite{he2016deep} structures to extract more robust pixel feature representations, a backbone network called HRNet was designed by Wang\cite{wang2020deep}, which maintains a high-resolution pixel representation during training. ResNet\cite{he2016deep} proposes a residual structure to refine the features at each layer, resulting in a more robust pixel representation. The other is to obtain more contextual information\cite{cao2019gcnet},\cite{fu2019dual},\cite{ruan2019devil},\cite{zhao2017pyramid},\cite{yang2018denseaspp},\cite{chen2017deeplab} to enhance the original pixel representations. The ways of capturing contextual information broadly include attention mechanisms\cite{fu2019dual}, using different dilation rates\cite{chen2018encoder} to increase the receptive field and building feature pyramids\cite{kirillov2019panoptic},\cite{liu2015parsenet}. In addition to this, the approach of taking care to balance the semantic features of classes when obtaining contextual information is proposed by\cite{jin2021isnet}. The focus of this paper is on the latter, i.e., enhancing the pixel representations by aggregating multi-distribution features per class.

\textbf{Context Aggregation}  Contextual information is extremely important for pixel-level classification tasks and is utilized in many research areas. The Deeplab\cite{chen2017deeplab},\cite{chen2017rethinking},\cite{chen2018encoder} family proposes multiple atrous convolution rates to increase the perceptual field of the network and thus aggregate more contextual information. DenseASPP\cite{yang2018denseaspp}, developed based on the Deeplab, stitches the dilated rates in a densely connected manner to generate multi-scale features that not only cover a wider range but are also dense. PSPNet\cite{zhao2017pyramid} adopts spatial pooling to obtain feature maps with different receptive field sizes. DANet\cite{fu2019dual} and OCNet\cite{yuan2018ocnet} calculate the pixel-to-pixel similarity as the weights, and weighted aggregation of all pixels to enhance the pixel representations. Apart from this, other methods\cite{li2019expectation},\cite{yuan2020object},\cite{zhang2019acfnet} proposed to divide pixels into multiple regions and enhance the feature representations of pixel by computing the similarity between pixels and regions as weights for weighted aggregation. Although the above methods are effective, they only extract the contextual information of a single image. But in this paper, our approach  learn class-level multiple distributions beyond input image to agument semantic informantion.

\textbf{Feature Memory Bank}  The feature memory bank can store the effective class features during the neural network training.  It has shown effectiveness in several computer vision work\cite{chen2020memory},\cite{li2019memory},\cite{wang2021exploring},\cite{alonso2021semi},\cite{wang2022semi},\cite{jin2022mcibi++},\cite{jin2021mining}. For instance, \cite{wang2021exploring} explores the contrastive loss of pixels between different images by saving the class features of the pixels through the memory bank in several batches. \cite{wang2022semi} use the memory bank to save the features of negative samples of each class to make the features of the same class similar and the features of different classes more discriminative. The memory bank of the above methods store the class-level features generated by bcakbone. \cite{jin2021mining}  uses  feature memory bank to store the class-level features that generated by simulating annealing. \cite{jin2022mcibi++} assumes that the class-level features satisfy a Gaussian distribution and store the mean and variance of each class by memory bank. In this paper, memory bank  is utilized by us to  store class-level multiple distribution features.

%\begin{figure*}
  % \centering
  % \includegraphics[width=\linewidth,height=8.7cm]{2}
 %  \caption{ \textbf{Overview of our method.} The pixel features generated by the backbone network are clustered to update the distribution in the dataset-level memory bank, while the features generated by the backbone network are enriched by class local consistency loss and class global consistency loss. The local dataset-level feature representation generated by voting are manipulated to obtain a global dataset-level feature representation. Finally refine the global dataset-level representation with an attention. Single image semantic aggregation is an optional operation that is specific to some existing methods (e.g., PPM[12],ASPP[2] ).}
%\end{figure*}

\begin{figure*}[t]
 \centering
  \includegraphics[width=\linewidth,height=8.7cm]{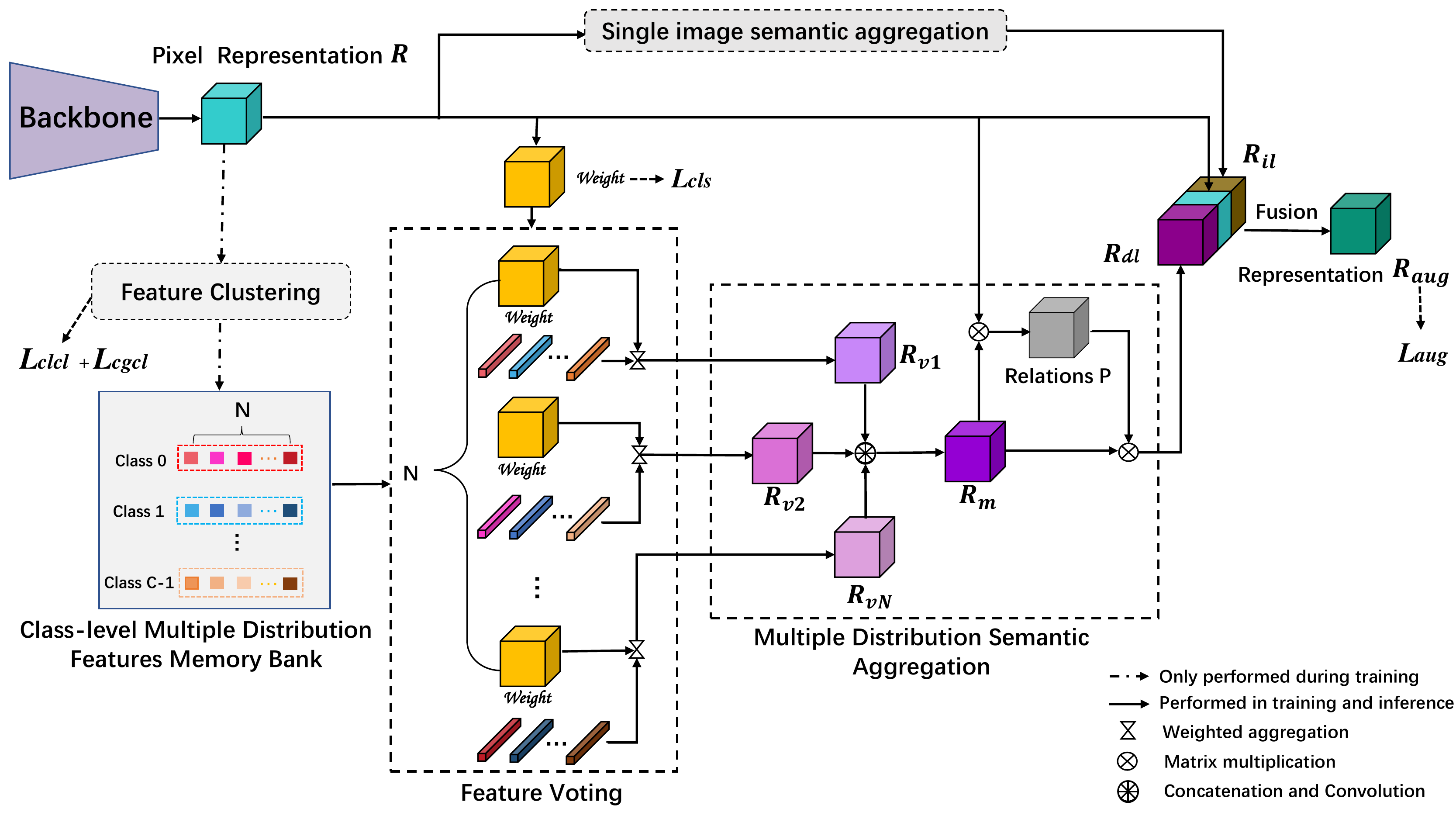}
  \caption{\textrm{\textbf{Overview of multiple distributions representation learning.}  The pixel representations $R$ are clustered by the class multiple distributions consistency strategy to update the class-level  multiple distribution features stored in   memory bank.  The single distribution  feature representations $R_{vi}$ generated by feature voting are aggregated by multiple distributions semantic aggregation module to obtain  a fine-grained multiple distribution feature representation $R_{dl}$. Single image semantic aggregation is an optional operation that is specific to some existing methods (e.g., PPM\cite{zhao2017pyramid},ASPP\cite{chen2017deeplab} ).}}
  \label{fig:3}
\end{figure*}

\section{Methodology}
In this section, we first introduce the overall flow of our method. Then describe the various parts involved in the method, including feature clustering, feature voting, multiple distributions semantic aggregation. Finally, we describe the details of class local consistency loss and class global consistency loss.

\subsection{Overview}
 Input an RGB image  $I \in {\mathbb{R}^{3 \times H \times W}}$,  mapping pixels to a non-linear embedding space through the backbone network ${N_t}$ as follows:
\begin{equation}\label{eq1}
  R{\rm{ = }}{N_t}(I)
\end{equation}
where the matrix $R$ denotes the pixel representations of $ I$ with the size  $Z \times \frac{H}{{\rm{8}}} \times \frac{W}{8}$. $Z$ stands for the number of channels. As shown in Figure \ref{fig:3}, we generate a more fine-grained multiple distribution feature representation ${R_{dl}}$ through the multiple distributions  semantic aggregation module:
\begin{equation}\label{eq2}
{R_{dl}} = DSA(R,\sum\limits_{i = 1}^N {fv(W{\rm{eg}},{m_i})} )
\end{equation}
where ${m_i}$ represents the $i$-th distribution feature of each class. $W{\rm{eg}}$ represents the class probability obtained after convolution and softmax of the feature map $R$. $fv()$ indicates the feature voting operation. $N$ indicates the number of distributions of each class in the memory bank. $DSA()$ represents multiple distributions semantic aggregation operation. The size of ${R_{dl}}$ is $Z \times \frac{H}{{\rm{8}}} \times \frac{W}{8}$.

Some methods\cite{zhao2017pyramid},\cite{chen2017deeplab} have the ability to aggregate the semantic information of  single image.  In order to  easily integrate our method into existing method frameworks, we refer to the single image semantic aggregation as ${SSA}$. We can obtain the semantic features of  single image:
\begin{equation}\label{eq3}
{R_{il}} = SSA(R)
\end{equation}
where ${R_{il}}$ is the semantic information of the current image. Next, the pixel representation $R$ are enhanced with multiple distribution feature representation ${R_{dl}}$ and single image semantic feature representation ${R_{il}}$ as follows:
\begin{equation}\label{eq4}
{R_{aug}} = T(R,{R_{il}},{R_{dl}})
\end{equation}
$T()$ is transform operator to fuse the basic representation $R$, ${R_{il}}$ and ${R_{dl}}$. Here it is important to note that  ${R_{il}}$ is optional. Finally, we need to classify the enhanced feature representation ${R_{aug}}$ and upsample it to obtain the corresponding semantic segmentation probability distribution map:
\begin{equation}\label{eq5}
O = U{P_{8 \times }}{\rm{(}}cls{\rm{(}}{R_{aug}}{\rm{))}}
\end{equation}
where $cls()$ is a classification head and $O$ is a matrix of $C \times H \times W$. $C$ is the number of classes in the dataset.

\subsection{Feature Clustering}
 The purpose of clustering is to bring pixel features of the same class closer together and those of different classes further apart. Clustering creates a natural bottleneck\cite{ji2019invariant} which  discard the details of instance-specific features. Therefore, we choose multi-distribution features of each class as subcenter of the corresponding class. 

Given the current batch of pixel ${P^c} = {\rm{\{ }}{{\rm{p}}_m}{\rm{\} }}_{m = 1}^M$ that belong to class c. $M$ denotes the number of pixels. Our goal is to aggregate these pixels of the same class into $N$ distributions  $\{ {d_{c,n}}\} _{n = 1}^N$. We denote pixel-to-distribution  mapping as ${L^c} = [{l_{{p_m}}}]_{m = 1}^M \in {\{ 0,1\} ^{N \times M}}$, where  ${l_{{p_m}}} = [{l_{{p_m},n}}]_{n = 1}^N \in {\{ 0,1\} ^N}$ is the one-hot assignment vector of pixels ${p_m}$  over the $N$ distributions.  Maximize the similarity between pixel embedding ${I^c} = [{p_m}]_{m = 1}^M \in {\mathbb{R}^{G \times M}}$ and distributions ${D^c} = [{d_{c,n}}]_{n = 1}^N \in {\mathbb{R}^{G \times N}}$ to optimize ${L^c}$
\begin{equation}\label{eq6}
\begin{array}{l}
\mathop {\max }\limits_{\quad \quad {L^c}} Tr({L^{c \top }}{D^{c \top }}{I^c}) + \lambda \sum\nolimits_{m,n} { - {l_{{p_m},n}}\log \;{l_{{p_m},n}}} \\
s.t.{\kern 1pt} \quad {L^c} \in _  \mathbb{R_+}^{N \times M},\;{L^{c \top }}{1^N} = {1^M}\;,\;{L^c}{1^M} = \frac{M}{N}{1^N}
\end{array}
\end{equation}
where ${1^N}$ represents a vector of N dimensions all 1. ${L^{c \top }}{1^N} = {1^M}$ ensures that each pixel is assigned to a distribution.  ${L^c}{1^M} = \frac{M}{N}{1^N}$ enforces that each distribution is selected at least $\frac{M}{N}$ times in the current batch\cite{caron2020unsupervised}.  $\lambda  > 0$ is a parameter that controls the smoothness of distribution. The solution of Eq.6 can be given as\cite{cuturi2013sinkhorn}:
\begin{equation}\label{eq7}
{L^c} = diag(u)exp(\frac{{{D^{c \top }}{I^c}}}{\lambda })diag(v)
\end{equation}
where $u \in {\mathbb{R}^N}$ and  $v \in {\mathbb{R}^M}$ are renormalization vectors which computed by few steps of Sinkhorn-Knopp iteration\cite{cuturi2013sinkhorn}.

\textbf{Feature Memory Update}  For any distribution of each class in the  memory bank, we choose pixels that closest to this distribution to update during each training iteration by following way:
\begin{equation}\label{eq8}
{d_{c,n}} = \mu {d_{c,n}} + (1 - \mu ){b_{c,n}}
\end{equation}
where $ {b_{c,n}}$ indicates the ${\ell _2}$-normalized vector of the embedded training pixels, which are closest to ${d_{c,n}}$. Here it is necessary to ensure that the class of the distribution is the same as the class of the pixels.  $\mu  \in [0,1]$ is a momentum update factor. Following\cite{zhou2022rethinking}, $\mu$ is set to 0.999.

\subsection{Feature Voting}
Feature memory bank  stores the  class-level multiple distribution  features $
\{ \{ {d_{c,1}},{d_{c,2}}, \cdot  \cdot  \cdot ,{d_{c,N}}\} \} _{c= 1}^C$.  ${D_i} = \{ {d_{c,i}}\} _{c = 1}^C$ is the distrubution feature of each class used for feature voting in $i$-th group. The basic feature  representations $R$ is convolved and softmax operated to obtain the weight matrix $weight$. Feature Voting is achieved by the following way:
\begin{equation}\label{eq9}
 {R_{vi}} = resize(permute(weight) \otimes {D_i})
\end{equation}
where $permute(weight)$ is used to make $weight$ have size of $\frac{{HW}}{{64}} \times C$, with the size of $C \times Z$ for ${D_i}$. $ \otimes$ denotes for the  matrix multiplication.  $Z$ denotes the dimension of the distribution feature,  $resize()$ represents the conversion of the size to  $\frac{H}{{\rm{8}}} \times \frac{W}{8}{\rm{ \times Z}}$.

\subsection{Multiple Distribution Semantic Aggregation}
Several single distribution feature representations  ${\rm{\{ }}{R_{vi}}{\rm{\} }}_{{\rm{i = 1}}}^N$ are obtained after feature voting. We fuse these representations to obtain the coarse multiple distribution feature representation ${R_{m}}$.  Since the predicted result is obtained by $R$ alone, pixels may be misclassified. To address this problem, we calculate the relations between $R$ and ${R_{m}}$ so that we can obtain a position confidence weight to further refine ${R_{m}}$,  We calculate the relations $P$ by the following equation:
\begin{equation}\label{eq10}
P = soft\max (\frac{{{g_a}(permute(R)) \otimes {g_b}{{({R_{m}})}^T}}}{{\sqrt {\frac{C}{2}} }})
\end{equation}
where $permute()$ aims to change the size of $R$ to $\frac{{HW}}{{64}} \times C$. ${R_{m}}$ is refined by following equation:
\begin{equation}\label{eq11}
{R_{dl}} = rescale({g_c}(P \otimes {g_f}({R_{m}})))
\end{equation}
where  ${g_a},{g_b},{g_c}$ and ${g_f}$  change the number of channels per pixel, $rescale()$ denotes to let the output have size of  $Z \times \frac{H}{8} \times \frac{W}{8}$.

\subsection{Loss Function}
\textbf{Class local consistency loss}   We use $N$ distribution features $\{ {d_{c,n}}\} _{n = 1}^N$ to characterize each class ${\rm{c}} \in \{ 1,. \cdot  \cdot  \cdot ,C\}$.  In this way, the detailed features of each class can  be captured by our method, the prediction of each pixel $i \in I$ is realized by the following way:
\begin{equation}\label{eq10}
\mathop {{c_{i}}}\limits^ \wedge   = {c^{\rm{*}}}\quad {\rm{w}}ith\;({c^{\rm{*}}},{n^*}) = \mathop {\arg \min }\limits_{(c,n)} \{ \langle i,{d_{c,n}}\rangle \} _{c,n = 1}^{C,N}
\end{equation}
where $i \in { \mathbb{R}^Z}$  denotes the ${\ell _2}$-normalized embedding of pixel $i$. $\langle  \cdot , \cdot \rangle $ is the negative cosine similarity as distance measure, just as  $\langle i,d\rangle  =  - {i^ \top }d$. We define the class probability distribution of pixel $i$ as:
\begin{equation}\label{eq11}
p({\rm{c|}}i) = \frac{{\exp ( - {s_{i,c}})}}{{\sum\nolimits_{c{\rm{c = 1}}}^C {\exp ( - {s_{i,cc}})} }}
\end{equation}
where ${s_{i,c}} = \min \{ \langle i,{d_{c,n}}\rangle \} _{n = 1}^N$, represents the distance between the current pixel $i$ and the  nearest distribution chosen from the class to which it belongs.  Given the ground-truth class of the current pixel $i$,  ${c_i} \in \{ 1, \cdot  \cdot  \cdot ,C\}$,  The cross-entropy loss of the class local consistency:
\begin{equation}\label{eq12}
\begin{array}{l}
{L_{clcl}} =  - \log p({c_i}|i)\\
\quad \;\;\, =  - \log \frac{{\exp ( - {s_{i,{c_i}}})}}{{\exp ( - {s_{i,{c_i}}}){\rm{ + }}\sum\nolimits_{{\rm{cc}} \ne {c_i}} {\exp ( - {s_{i,cc}})} }}
\end{array}
\end{equation}
The operation of the  class local consistency loss lies in first filtering out a distribution feature of each class that is closest to the current pixel ${i}$.  Eq.\ref{eq12}   can be viewed as pushing  pixel $i$ closer to the nearest distribution feature of its corresponding class,  and far away from other  close distribution features of unrelated classess.  

\textbf{Class global consistency loss}  The class local consistency loss only utilize  one distribution of each class. With the idea of fully using multi-distribution features of each class, we design the class global consistency loss to fulfill intra-class compactness and  inter-class dispersion.  Ablation experiment proves that the class global consistency loss and the class local consistency loss complement each other to better improve the segmentation performance.
%During the experimental stage, the optimal result was not achieved by class local consistency loss alone. We guess there are 2 reasons for this phenomenon. The first reason is that the nearest distribution centers of the ground-truth class may not differ significantly from the nearest distribution centers of the other classes during the training process. The second reason is that there are too few distribution centers for each class used for class local consistency loss, which lacks representation from a certain extent. Therefore, we propose class global consistency loss to assist class local consistency loss for clustering.
%

For pixel $i$, class global consistency loss takes ${N}$ distribution features of the true class as positive samples and $(C - 1) \times N$ distribution features of other unrelated classes as negative samples for contrastive loss. We define the class global consistency loss as:
\begin{equation}\label{eq13}
\begin{array}{l}
{L_{cgcl}} =  - \log \frac{A}{{{\rm{A + B}}}}\\
A = \sum\nolimits_{j = 1}^N {\exp ({{ - \langle {z_{ci}},z_{ci,j}^ + \rangle } \mathord{\left/
 {\vphantom {{ - \langle {z_{ci}},z_{ci,j}^ + \rangle } \tau }} \right.
 \kern-\nulldelimiterspace} \tau })} \\
B = \sum\nolimits_{jj = 1}^{(C - 1) \times N} {\exp ({{ - \langle {z_{ci}},z_{ci,jj}^ - \rangle } \mathord{\left/
 {\vphantom {{ - \langle {z_{ci}},z_{ci,jj}^ - \rangle } \tau }} \right.
 \kern-\nulldelimiterspace} \tau })} 
\end{array}
\end{equation}
where ${z_{ci}}$ denotes the feature of pixel ${i}$, $z_{ci,j}^ +$ represents the positive sample features of pixel $ {i} $, $z_{ci,jj}^ - $ denotes the negative sample features of pixel ${i}$.  Following \cite{liu2021bootstrapping}, the ${\tau }$ is set to 0.5. 

\textbf{Classification loss}  There are two types of classification losses in our approach. One is for the loss of the basic feature representation ${R}$ as:
\begin{equation}\label{eq14}
{L_{cls}} = \frac{1}{{H \times W}}\sum\limits_{(i,j)} {{L_{{\rm{ce}}}}(trans({R_{[*,i,j]}}),G{T_{[ij]}})} 
\end{equation}
and the other is for the loss of the enhanced feature representation  ${{R_{aug}}}$ as:
\begin{equation}\label{eq15}
{L_{aug}} = \frac{1}{{H \times W}}\sum\limits_{(i,j)} {{L_{{\rm{ce}}}}(trans({R_{aug}}_{[*,i,j]}),G{T_{[ij]}})} 
\end{equation}
where ${L_{{\rm{ce}}}}$ represents the cross entropy loss, $trans()$ denotes the transformation of the feature representation into a class probability distribution map.  $GT$ represents the ground-truth.

We optimize the overall loss as follows:
\begin{equation}\label{eq15}
L = \eta {L_{cls}} + {L_{aug}} + \alpha {L_{clcl}} + \beta {L_{cgcl}}
\end{equation}
Following \cite{jin2021mining}, we set $\eta $ as 0.4. We do detailed ablation experiments for weight $\alpha $ and $\beta$ of class local consistency loss and class global consistency loss respectively. We obtained the optimal weight after the ablation experiment, which set  $\alpha  = 0.01$ and $\beta = 0.05$.

\section{Experiments}
\subsection{Experimental Setup}
\textbf{Datasets} We conduct experiments on two widely-used semantic segmentation benchmarks:

$\bullet \textbf{Cityscapes}$\cite{cordts2016cityscapes}  is derived from urban scene. It has 5,000 finely annotated data with 19 classes. The dataset is divided into 2975/500/1525 for train/val/test.

$\bullet \textbf{ADE20K}$\cite{zhou2017scene} is a large scene parsing dataset that includes 150 classes. The training, validation and test sets consists of 20K,2K,3K images, respectively.

\textbf{Training Settings}  Our experiments are based on the PyTorch framework. The backbone networks adopt ResNet50 and ResNet101\cite{he2016deep} which pretrained on ImageNet\cite{deng2009imagenet}. Color jitter, random scaling and horizontal flipping are used for data augmentation. We adopt SGD algorithm to optimize the network parameters, learning rate is updated by poly strategy with factor ${(1{\rm{ - }}\frac{{{\rm{iter}}}}{{\max \_iter}})^{0.9}}$. The detailed training settings on each dataset are listed as follows:

\textbf{Cityscapes:}  The initial learning rate adjusted as 0.01, with weight decay is set to 0.0005. The input size of the image for the neural network is 512×1024, training epochs and batch size are set as 220 and 8, respectively.

\textbf{ADE20K:} We set the initial learning rate as 0.01, with weight decay is set to 0.0005. We crop the image size to 512×512 as the input to the network, batch size as 16 and training epochs as 130.

\textbf{Inference Settings} For the inference, we set the batch size to 1, the input image size is the same as the original image, but note that we also need multi-distribution features of each class for enhanced basic representations  during the inference.

\textbf{Evaluation Metrics} The standard mean intersection-over-union(mIoU) is used by us to measure the performance of the algorithm.

\textbf{Reproducibility}  Our approach is based on pytorch( version$=$1.10.0), trained on eight NVIDIA 3090 GPUs with a 24 GB memory per-card and two NVIDIA A40 GPUs with a 48 GB memory per-card. And all the testing procedures are performed on a NVIDIA A40 GPU.

\subsection{Ablation Study}
\noindent\textbf{Integrated into popular segmentation frameworks.}  As illustrated in Table \ref{tab:1}, we can see that our approach has greatly improved the performance of the popular network(FCN, PSPNet, CCNet). For instance, our approach brings 4${\%}$/5.61$\%$ mIoU improvements to FCN framework on Cityscapes/ADE20K.  And for stronger frameworks PSPNet/CCNet, the performance can get about 1.5${\%}$/0.63${\%}$  mIoU improvements on Cityscapes and 1.75${\%}$/0.75${\%}$ mIoU improvements on ADE20K. Experimental results demonstrate that our approach not only can integrates seamlessly into popular frameworks, but also delivers a high level of improvement.

\begin{table}[htbp]
\begin{center}
\begin{tabular}{|p{1.45cm}<{\centering} | c p{0.6cm}<{\centering}   p{1.3cm}<{\centering}   c|}
\hline
Method&Backbone&Stride&Cityscapes&ADE20K  \\
\hline\hline
 FCN\cite{long2015fully} & ResNet-50 & ×8 & 75.16 & 36.96 \\
 +ours & ResNet-50  &  ×8 & \textbf{79.16} & \textbf{42.57}  \\
PSPNet\cite{zhao2017pyramid}&ResNet-50&×8 &78.55&42.64\\
 +ours&ResNet-50&×8&\textbf{80.05}&\textbf{44.39} \\
 CCNet\cite{huang2019ccnet}&ResNet-50&×8&79.15&42.47 \\
+ours&ResNet-50&×8&\textbf{79.78}& \textbf{43.22}  \\
\hline
\end{tabular}
\end{center}
\caption{\textbf{The improvement when combining our approach with popular frameworks.} All the results are based on a single-scale validation. }
\label{tab:1}
\end{table}

\noindent\textbf{Number of class distributions}   As illustrated in Table \ref{tab:2}, different number of distributions can cause large different results. The divergence in performance between using 7 distributions and 9 distributions per class is close to 1.69$\%$ mIoU. Meanwhile, the results illustrate that it is necessary to use multiple distribution features to characterize the class representation, rather than just one. In other experiments, we used 9 distribution features to describe the intra-class variations .

\begin{table}[htbp]
\begin{center}
\begin{tabular}{|l|cccc|}
\hline
Method & Backbone &Dataset  &num & mIoU \\
\hline\hline
\multirow{8}{*}{FCN\cite{long2015fully}} &\multirow{8}{*}{ResNet-50}  & \multirow{8}{*}{Cityscapes}&1&78.50 \\ 
&  &  &3&78.64 \\  
&&    &5&78.66\\    
& &   & 7 & 77.47\\  
&  &&     9 & \textbf{79.16}\\ 
&&    &  11& 78.74\\  
&  &  &  13 & 78.35\\   
&  &  &15  & 78.54\\  
\hline
\end{tabular}
\end{center}
\caption{\textbf{Performance comparison under different distribution numbers.} Num represents the number of distributions per class. }
\label{tab:2}
\end{table}

\noindent\textbf{Validity of class local/global consistency loss}    We verified the two losses separately. As illustrated in Table \ref{tab:3}, using class local consistency loss alone can result in 3.04$\%$ mIoU improvements while class global consistency loss alone is 2.43$\%$ mIoU  improvements. It seems that class local consistency loss is more important than class global consistency loss, as seen by the results. But the simultaneous use of two losses can bring 4$\%$ mIoU performance improvements. This shows that the two losses complement each other and together improve the performance of the segmentation.

\noindent\textbf{Coefficient $\alpha$ and $\beta$}    We set several different sets of hyperparameters to compare the results. As illustrated in Table \ref{tab:4}, the performance of the model is more sensitive to the values of  $\alpha$  and $\beta$ . When $\alpha$ is set to 0.01, $\beta$ is set to 0.1, compare to 0.01, the difference in performance is approximately 0.9$\%$. According to the results, we obtain the optimal pair of parameters, which $\alpha$ is 0.01, $\beta$ is 0.05. Similarly, we set such optimal parameter pairs in other comparative experiments.

\begin{figure*}[t]
	\begin{tabular}{rccccc}
	FCN
      &\includegraphics[width=0.16\textwidth]{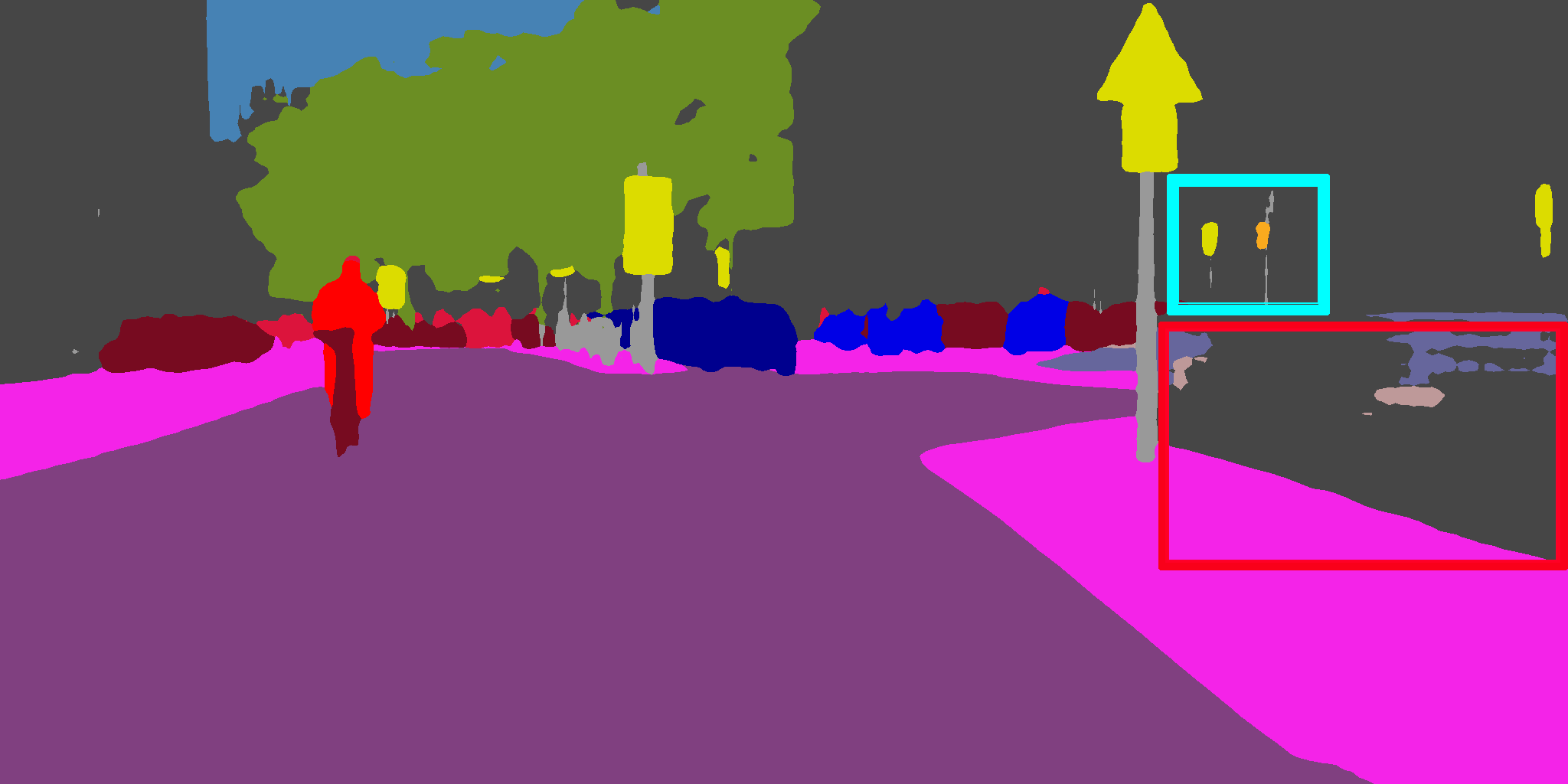}
      &\includegraphics[width=0.16\textwidth]{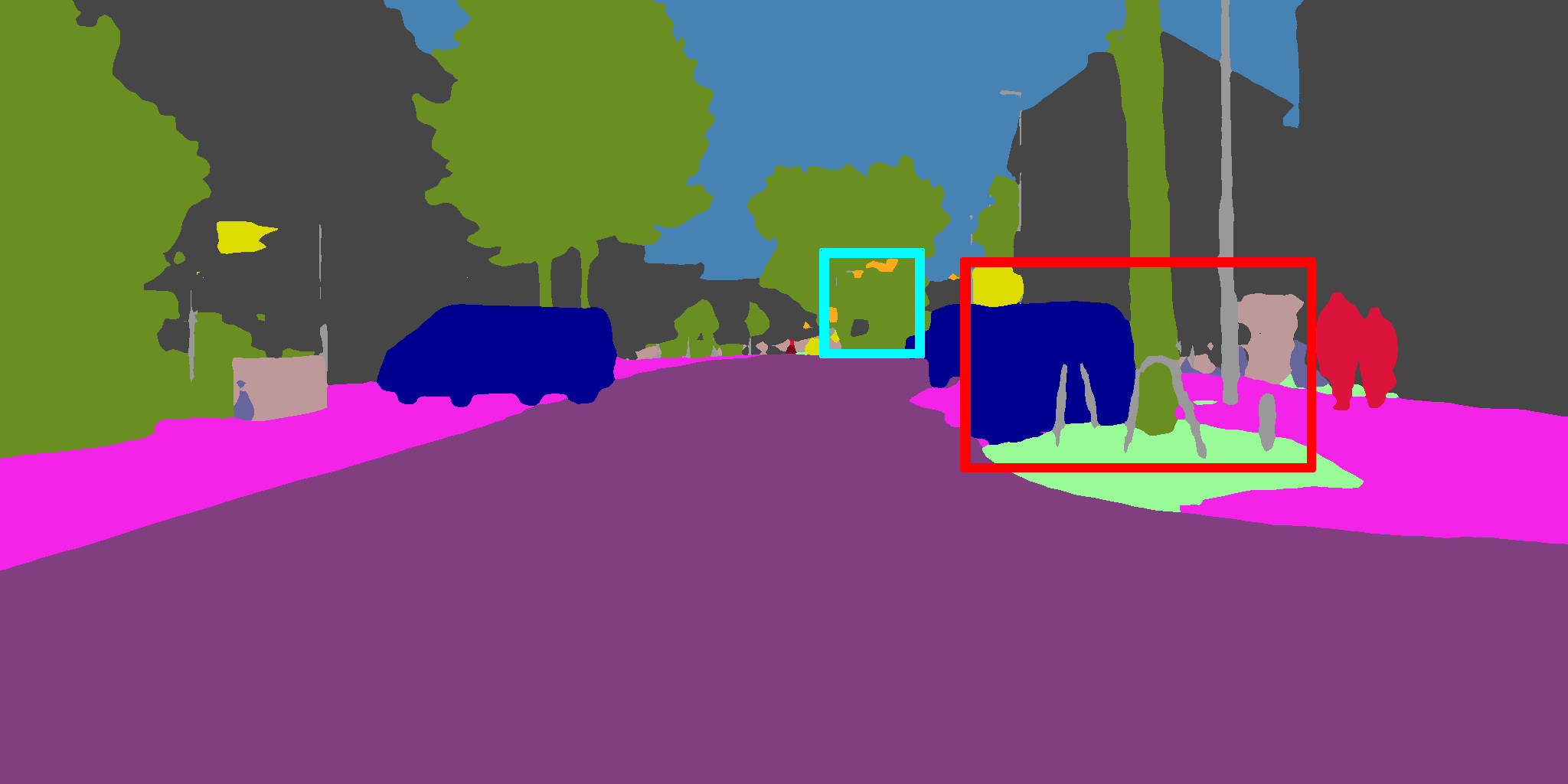}
      &\includegraphics[width=0.16\textwidth]{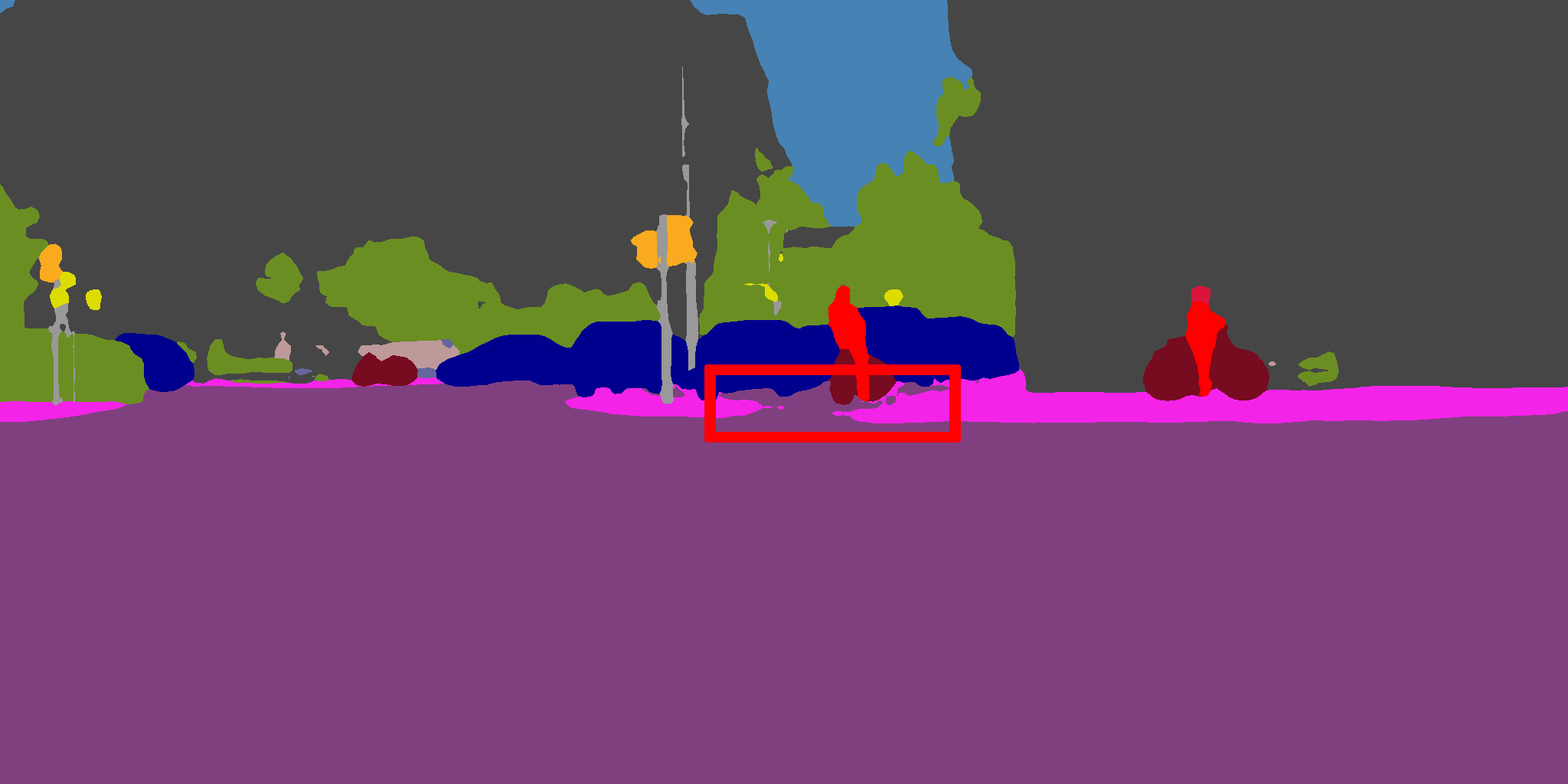}
      &\includegraphics[width=0.16\textwidth]{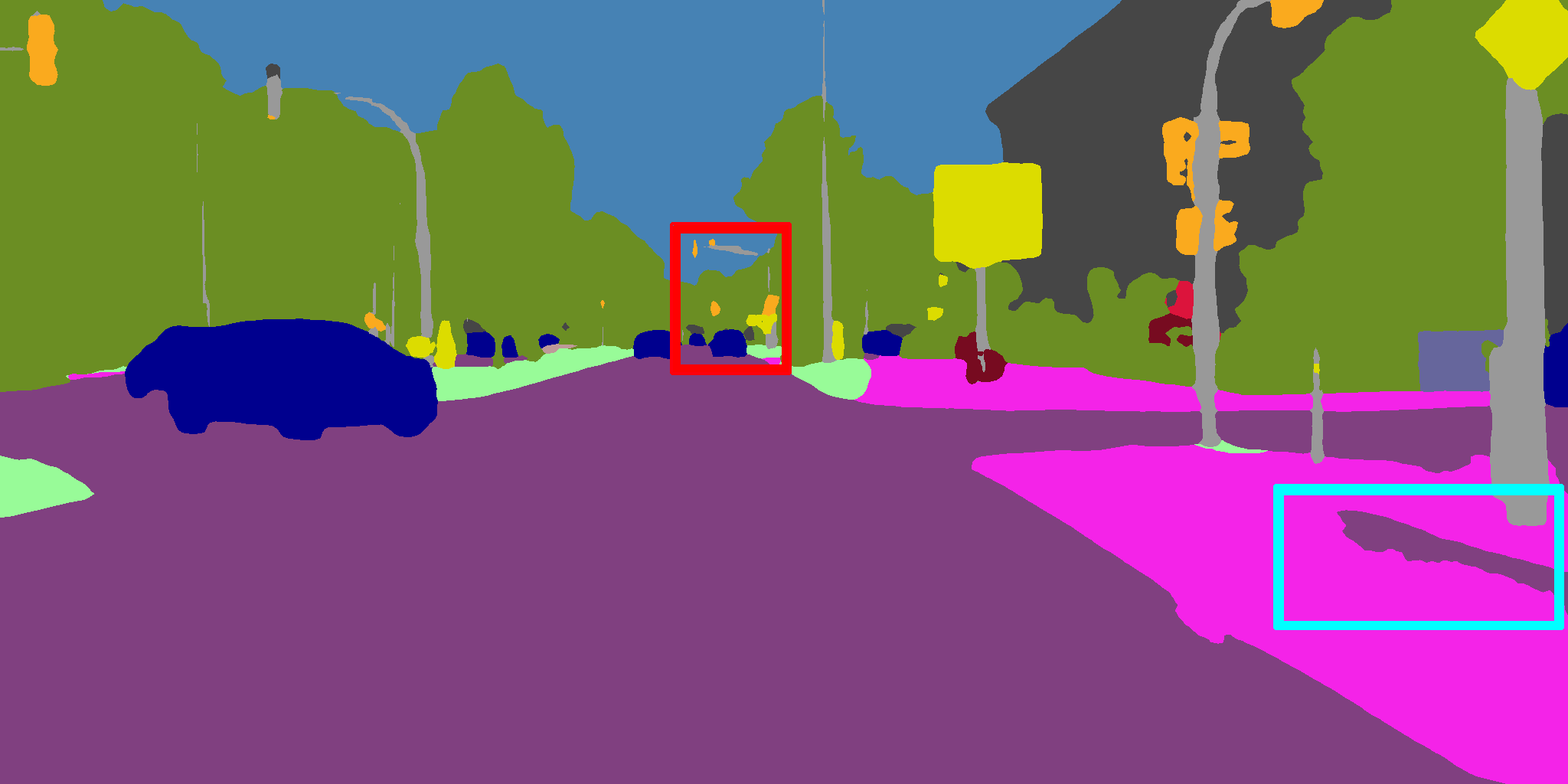}
      &\includegraphics[width=0.16\textwidth]{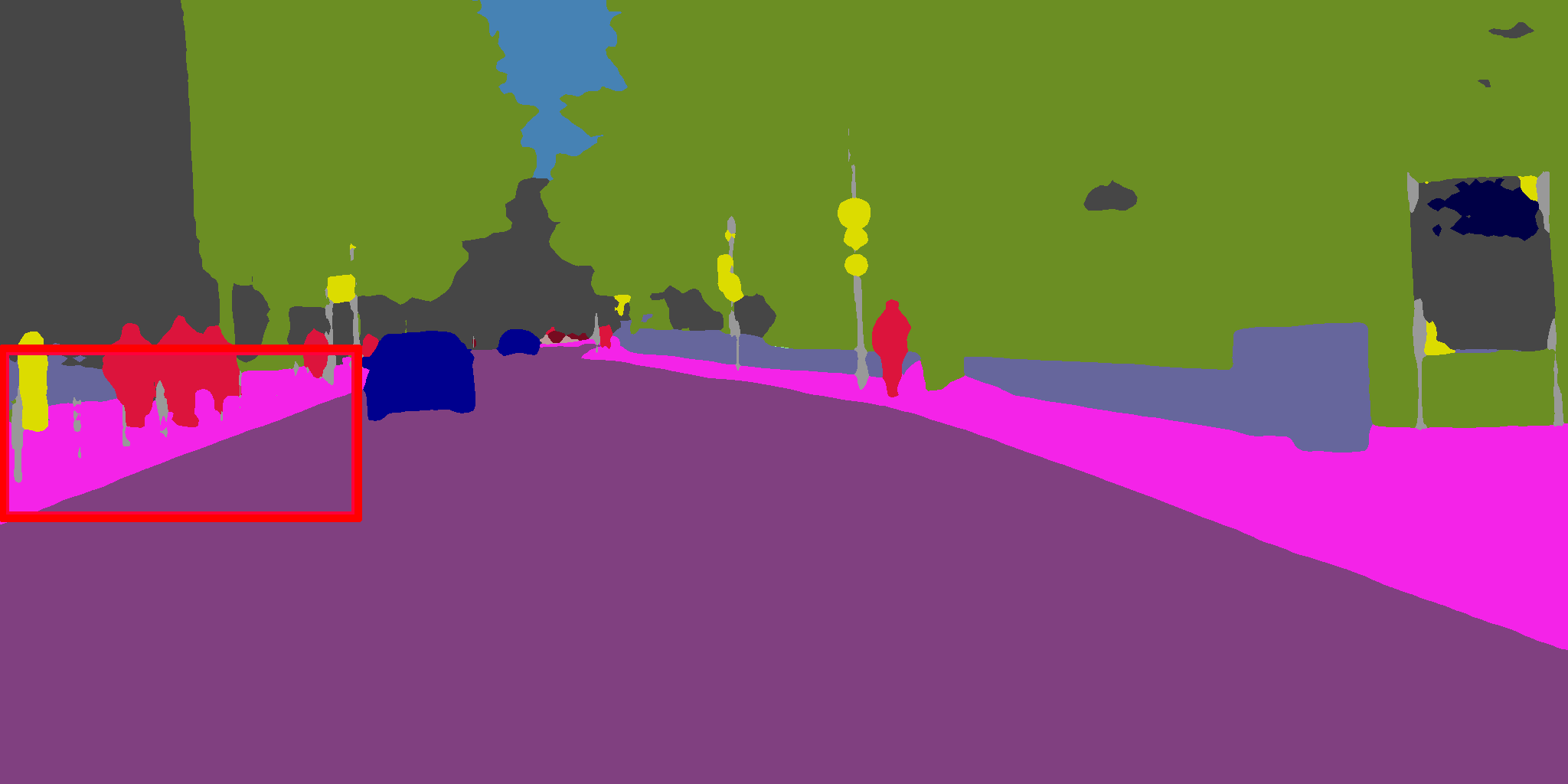}\\
       +our
      &\includegraphics[width=0.16\textwidth]{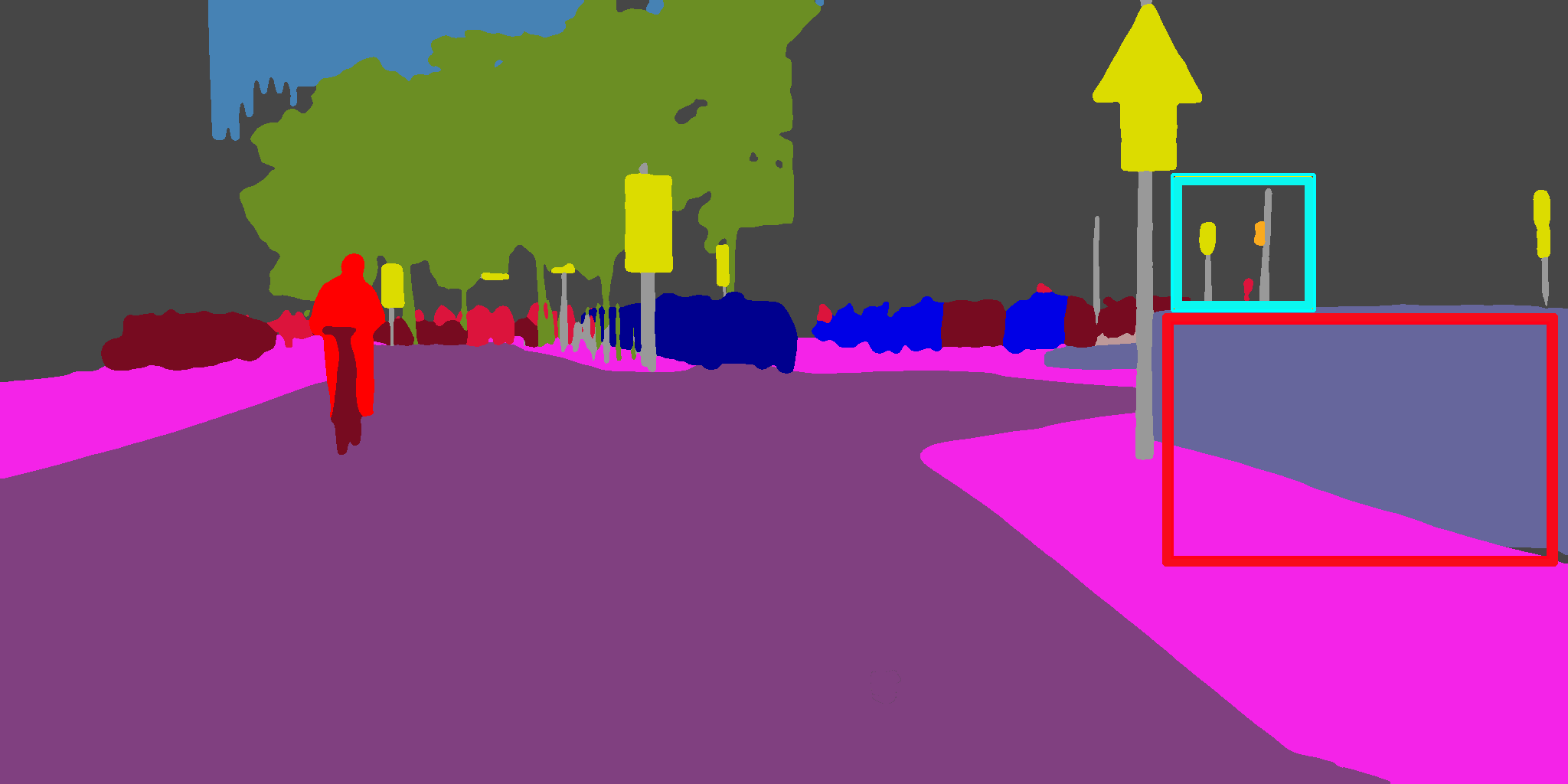}
      &\includegraphics[width=0.16\textwidth]{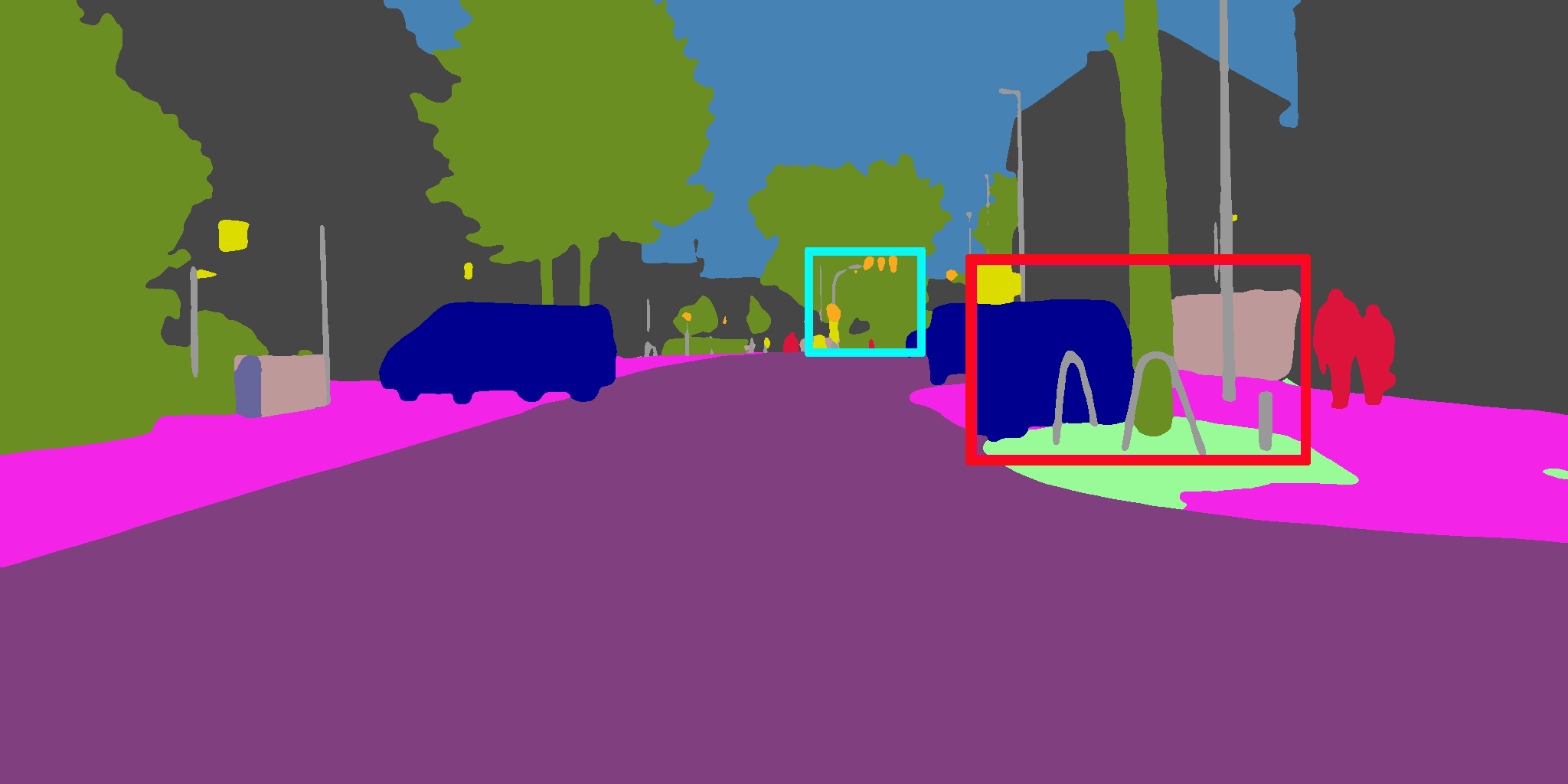}
     &\includegraphics[width=0.16\textwidth]{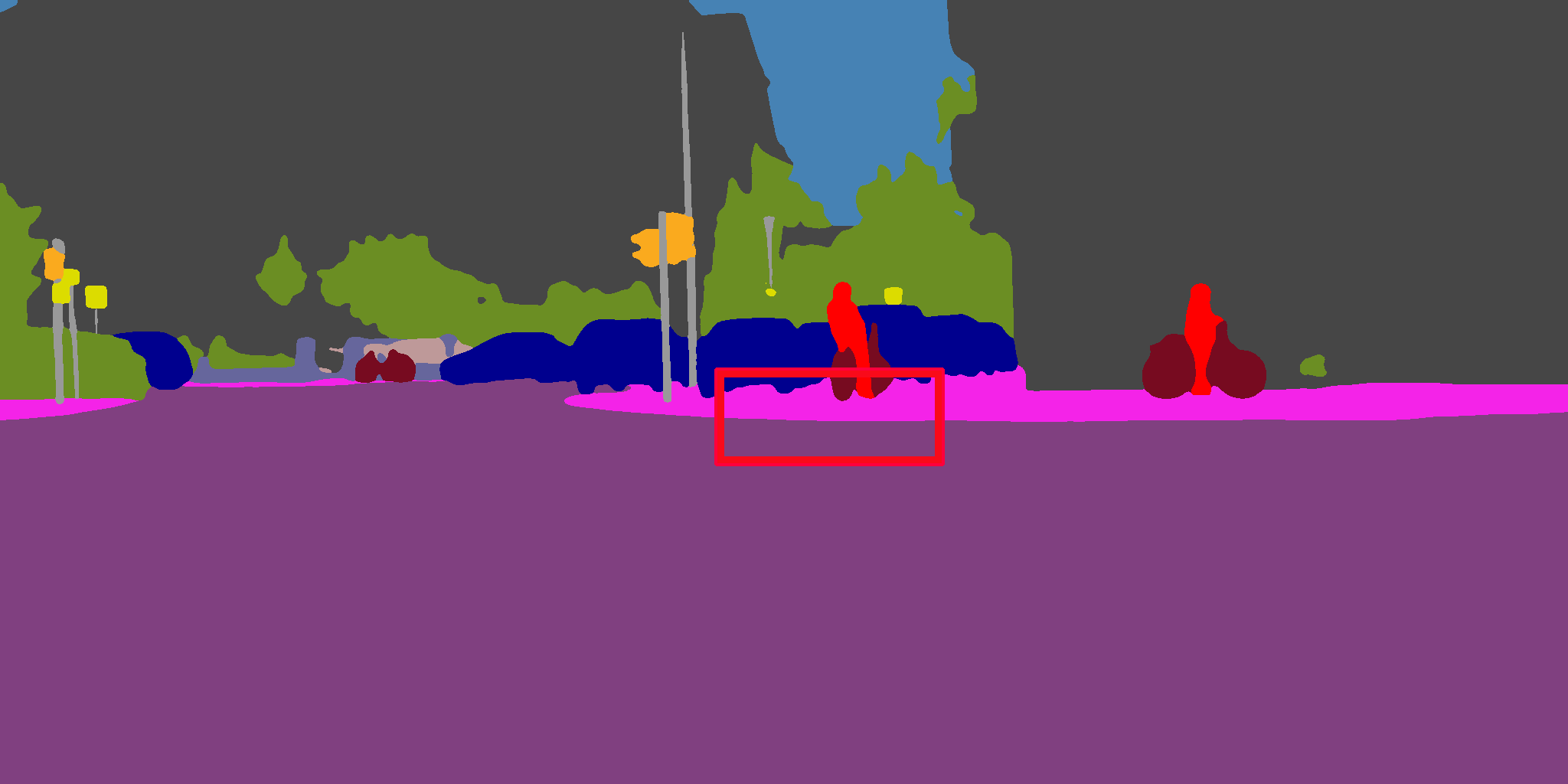}
      &\includegraphics[width=0.16\textwidth]{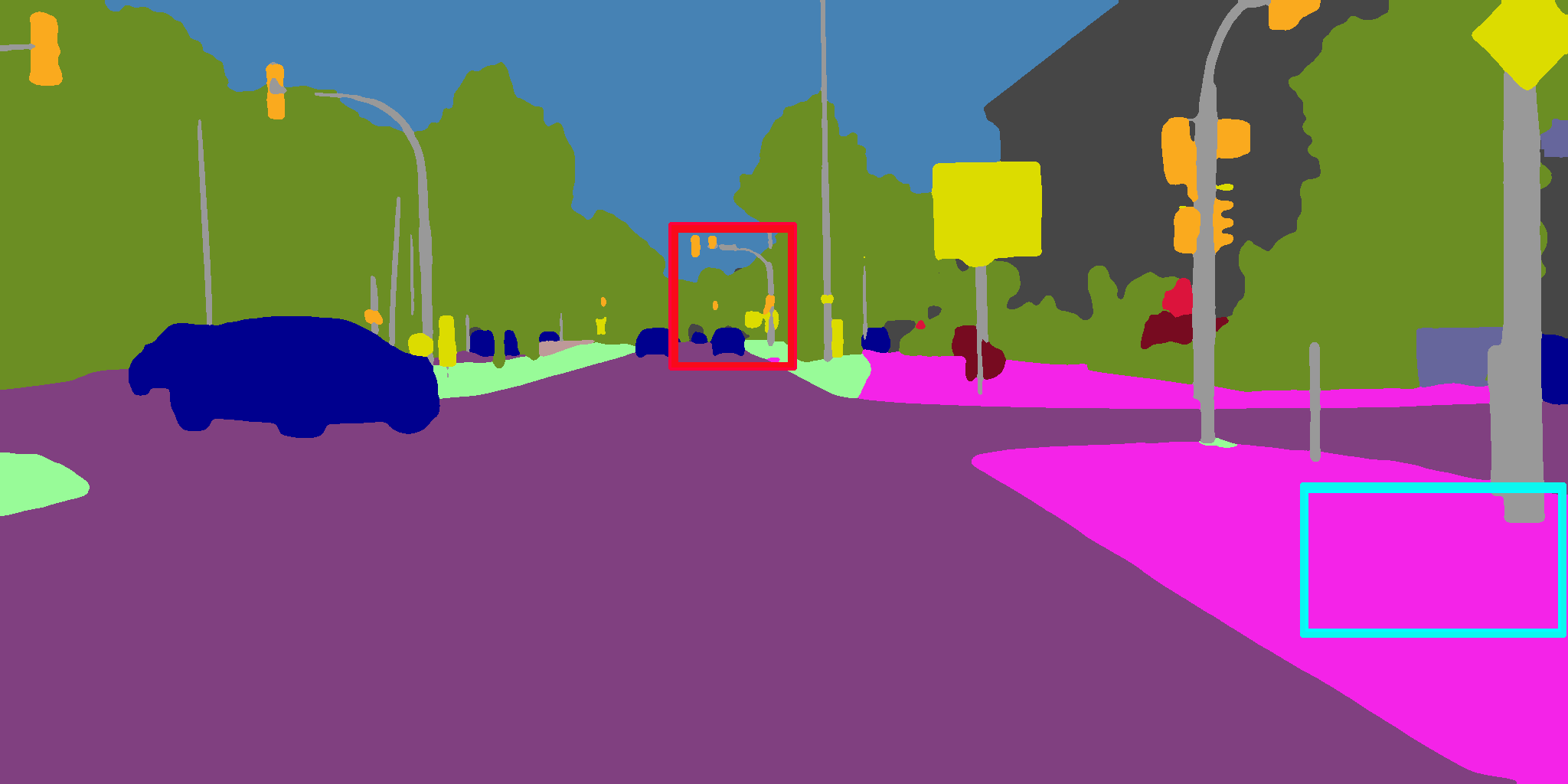}
     &\includegraphics[width=0.16\textwidth]{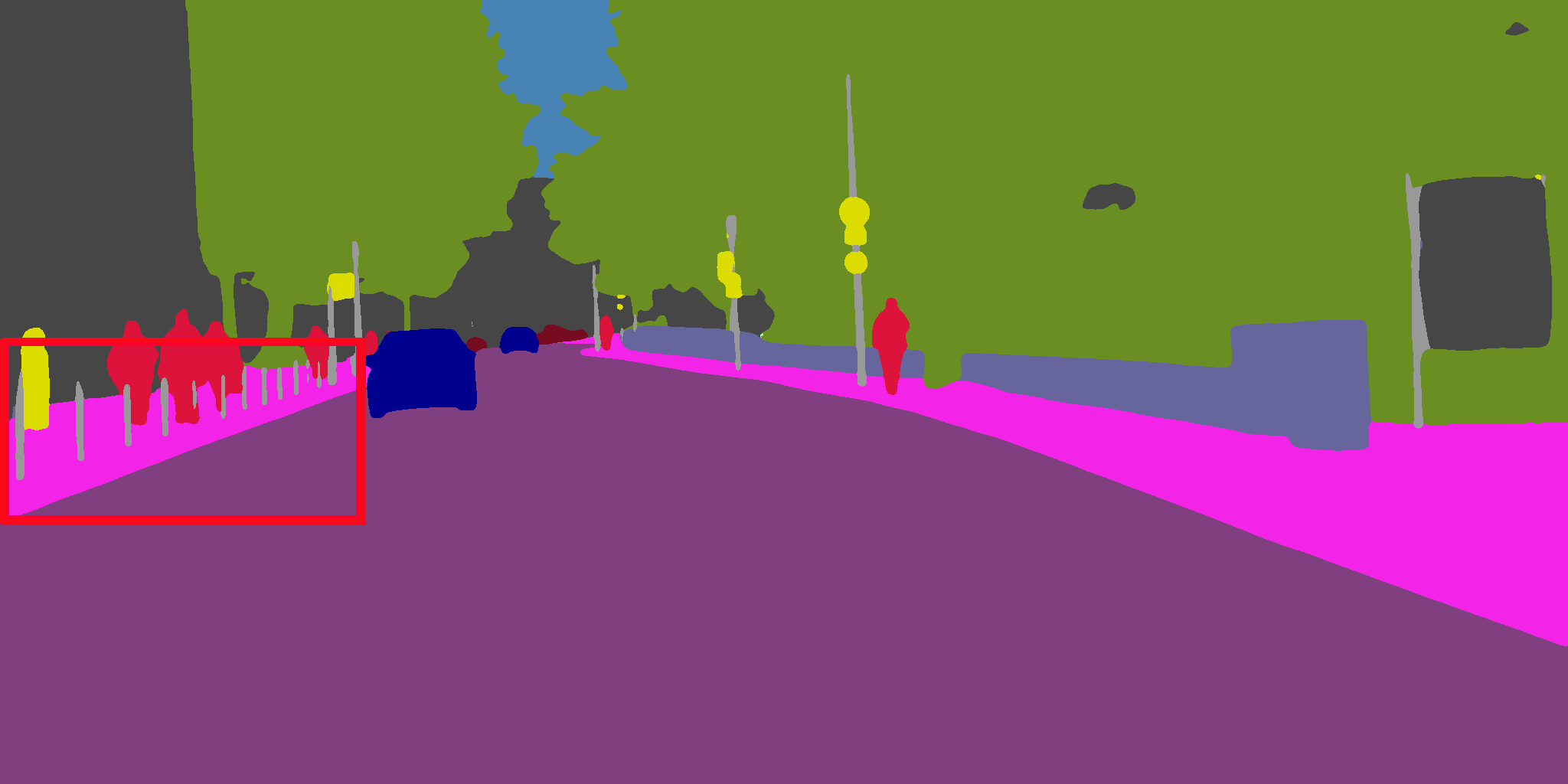}\\

      GT 
      &\includegraphics[width=0.16\textwidth]{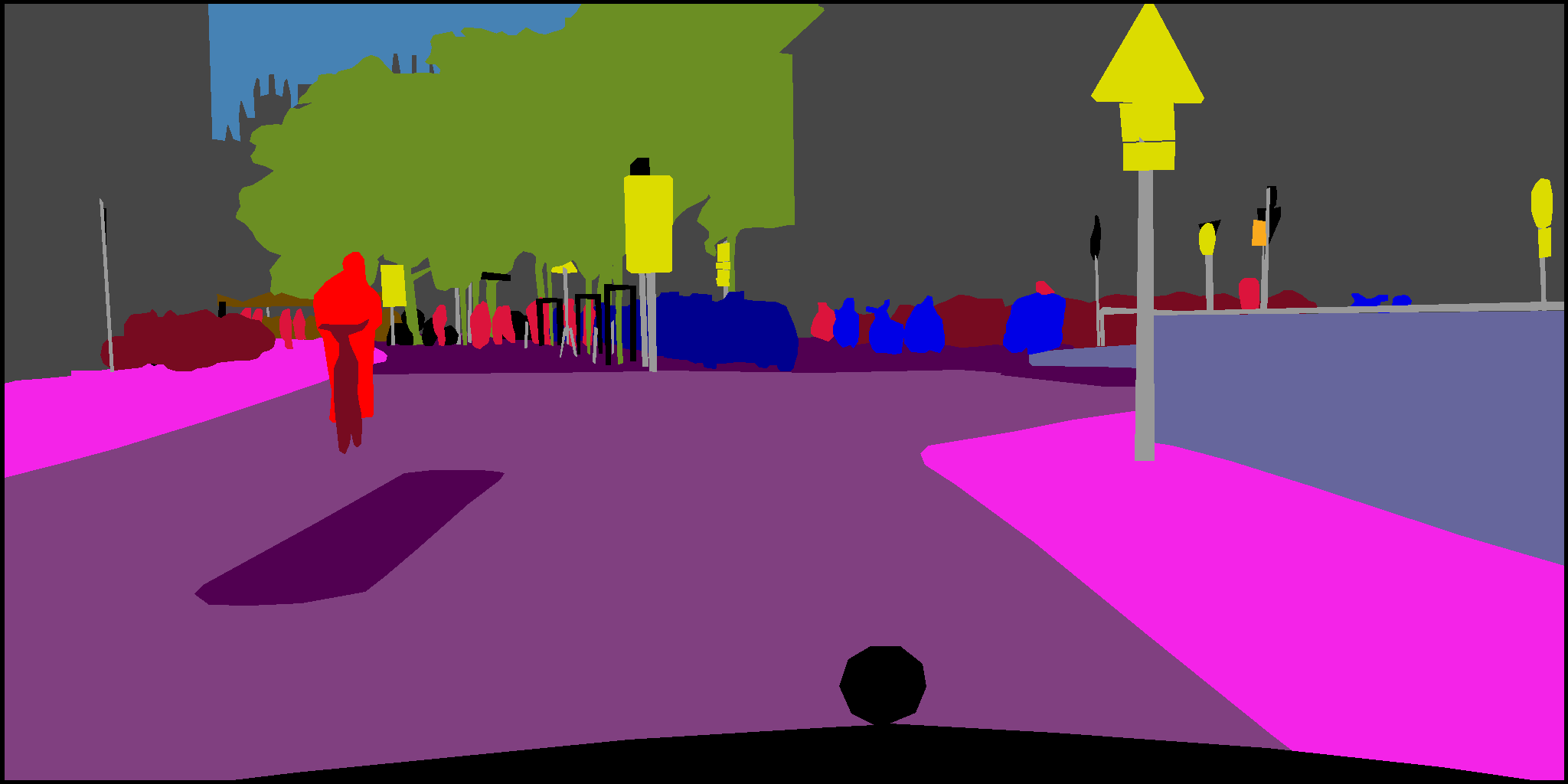}
      &\includegraphics[width=0.16\textwidth]{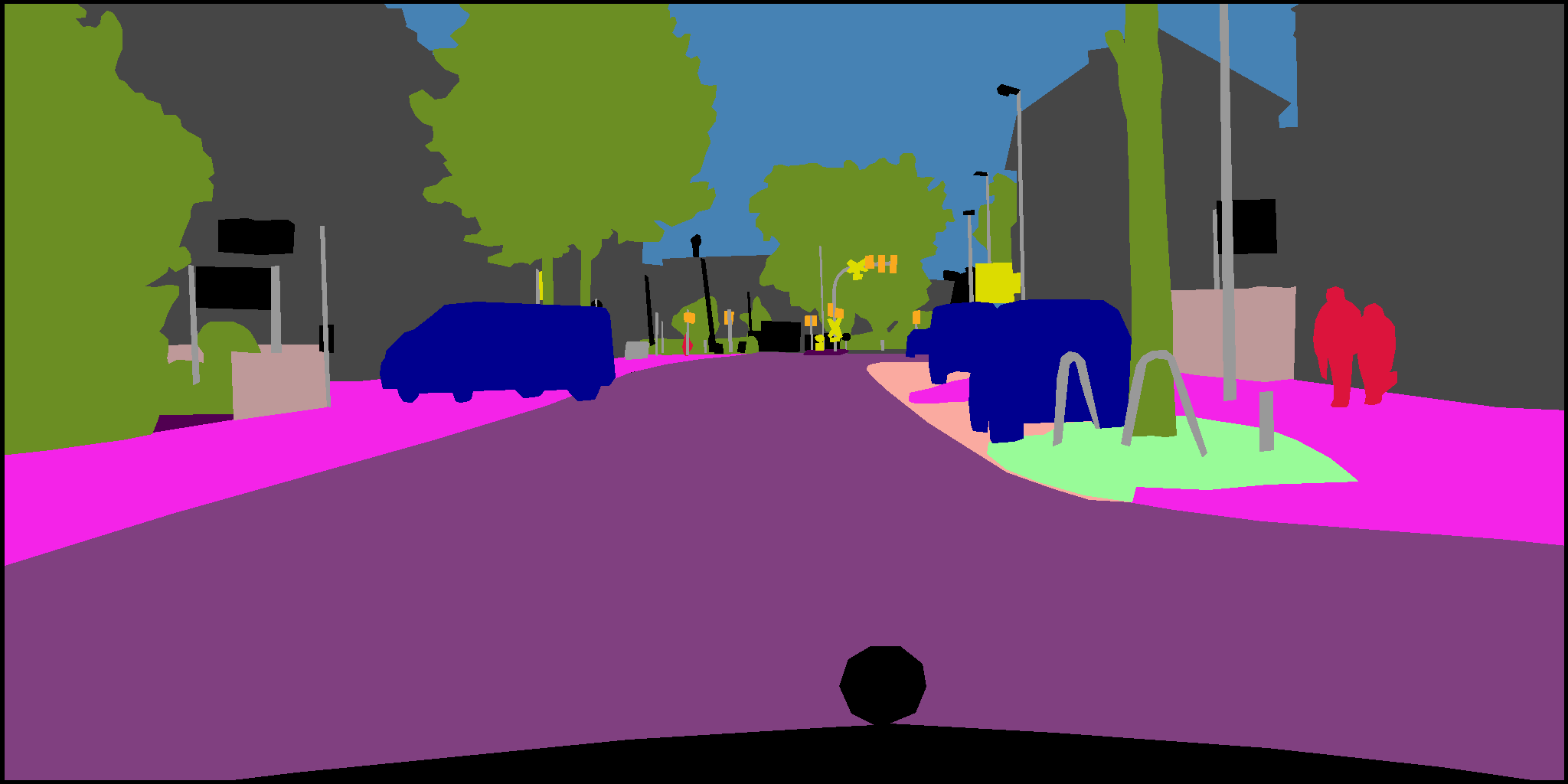}
      &\includegraphics[width=0.16\textwidth]{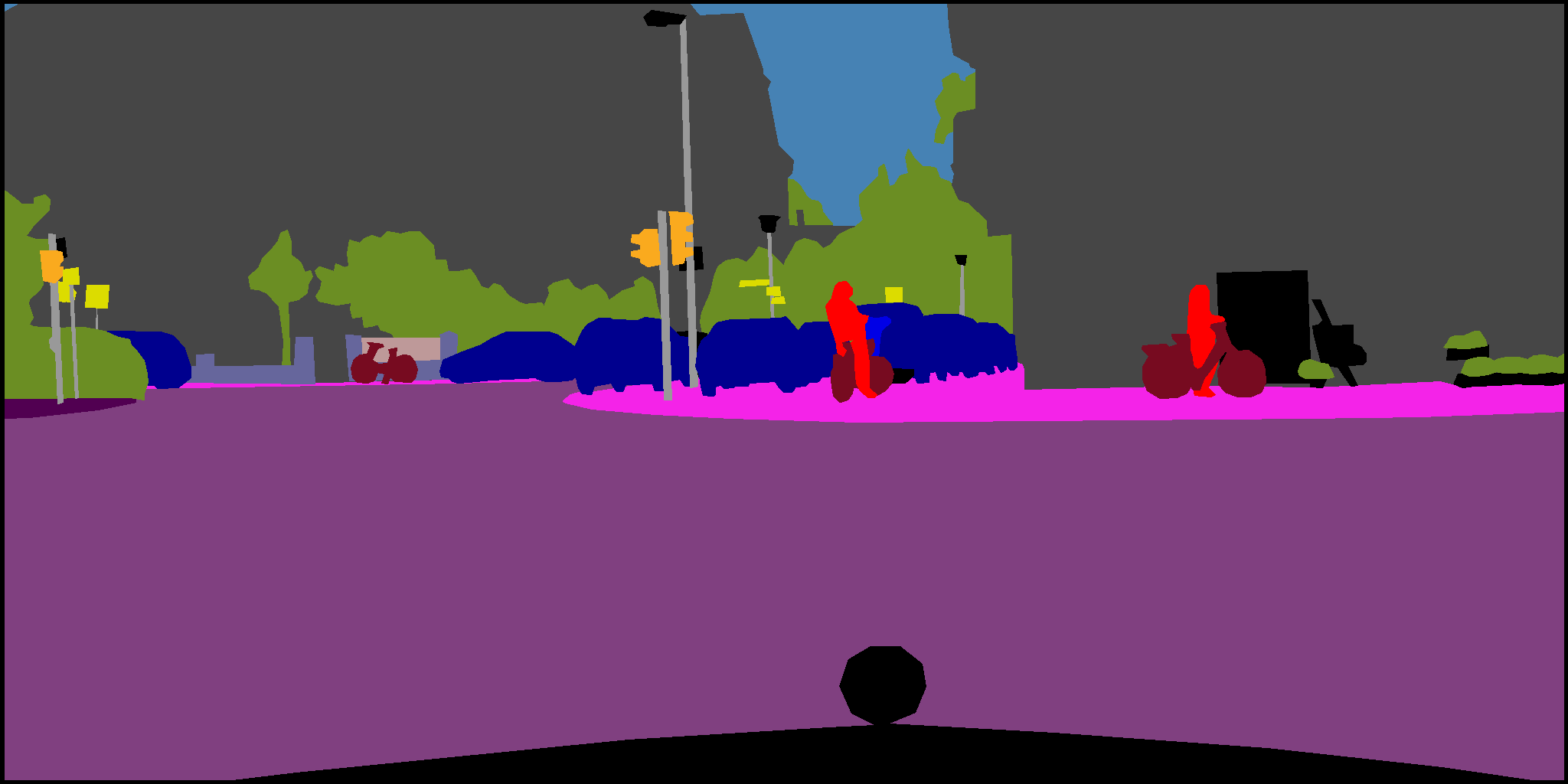}
      &\includegraphics[width=0.16\textwidth]{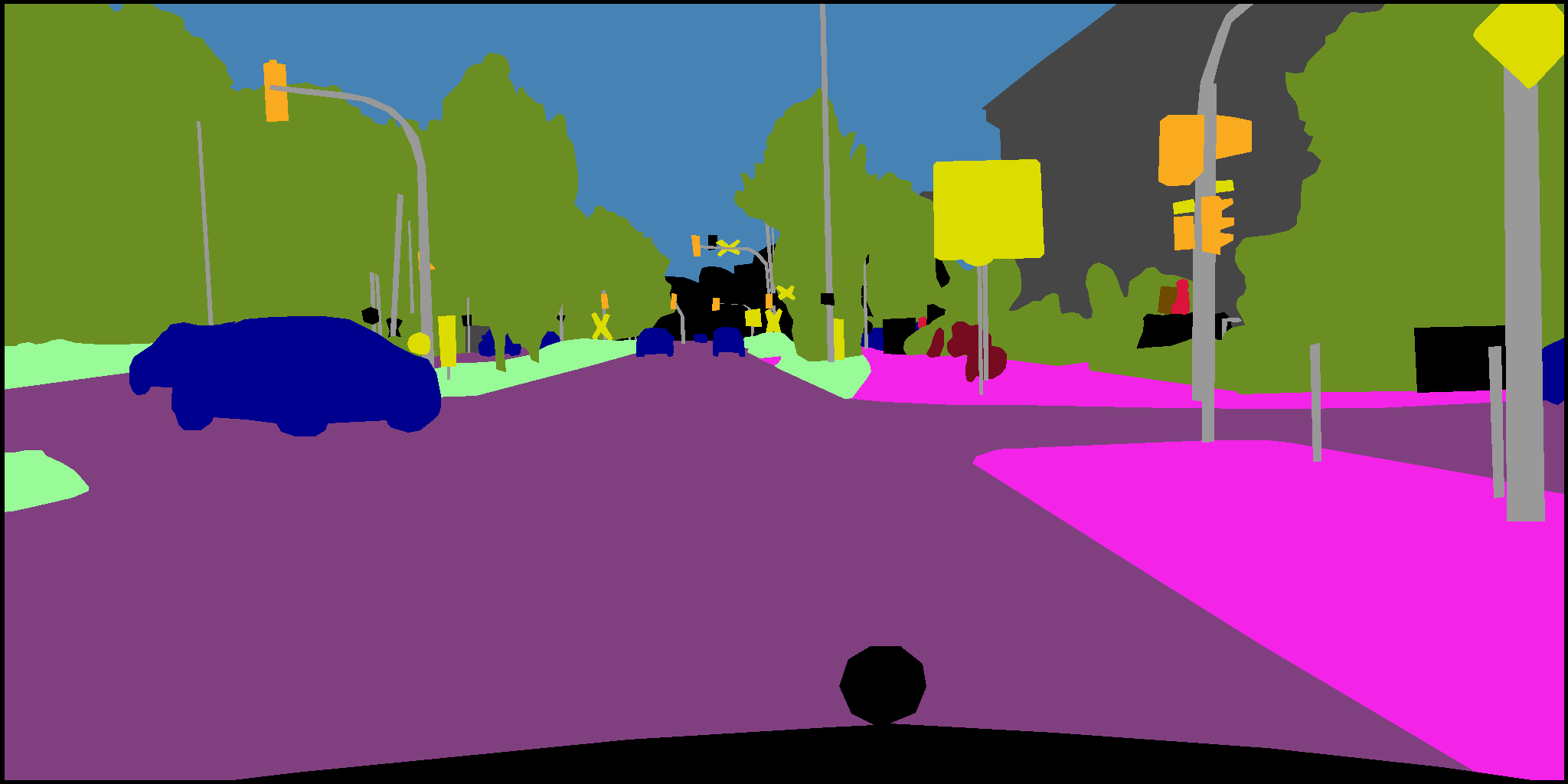}
     &\includegraphics[width=0.16\textwidth]{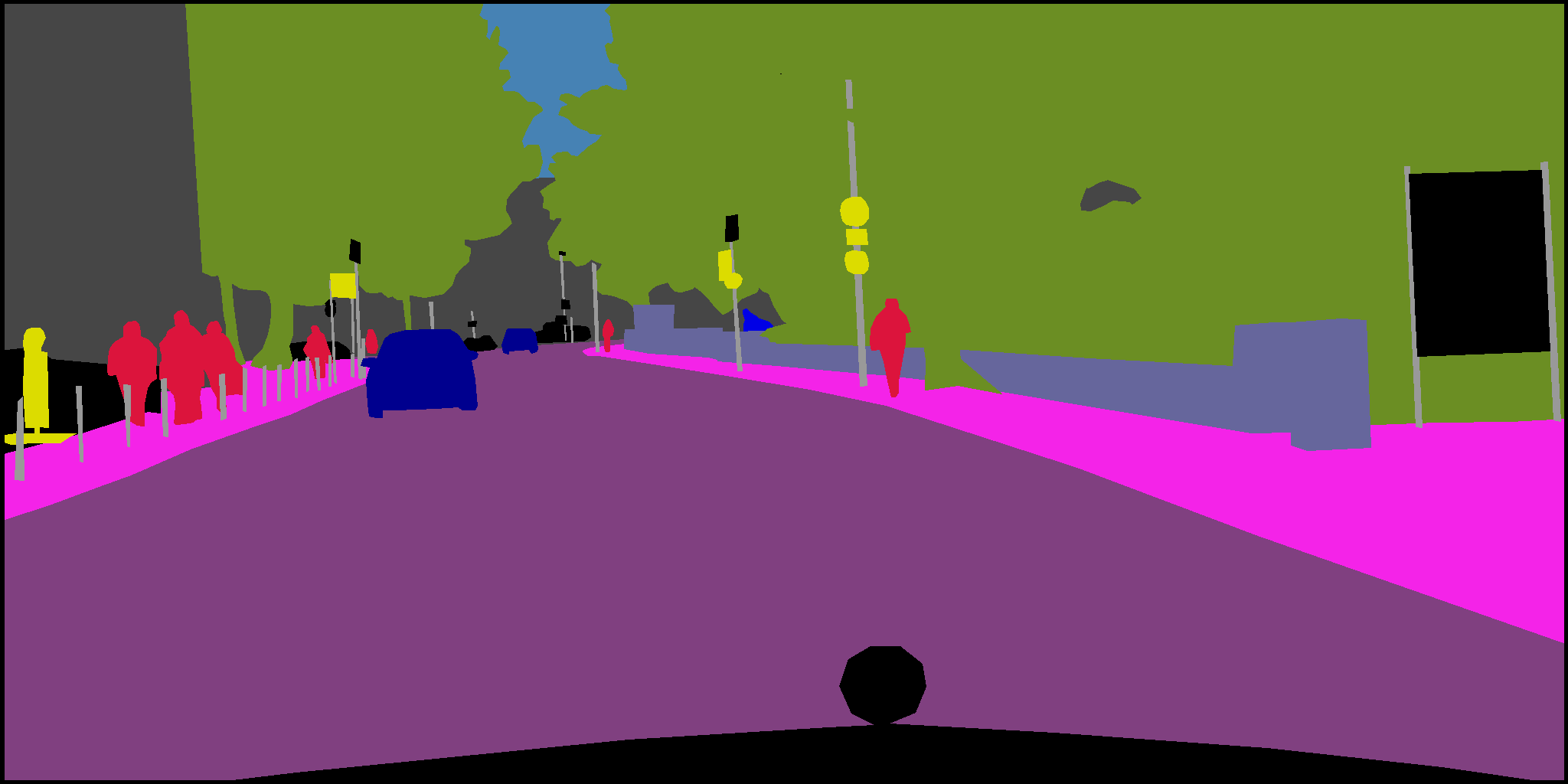}
	\end{tabular}

	\caption{\textbf{Visual comparsions with FCN and FCN+MDRL on the validation set of Cityscapes.}The last line indicates the ground-truth.}

	\label{fig:4}

\end{figure*}

\noindent\textbf{Visualization results with Baseline}  We integrated MDRL into the FCN network and obtained the segmentation results. As shown in Figure \ref{fig:4},compare with FCN, our method appears to be superior.

\begin{table}[htbp]
\begin{center}
\begin{tabular}{|c|ccc|}
\hline
 BaseLine& $L_{clcl}$& $L_{cgcl}$&mIoU \\
\hline\hline
 ${\checkmark}$& & &  75.16 \\ 
  $\checkmark$&$\checkmark$& & 78.20   \\  
 $\checkmark$& &$\checkmark$& 77.59  \\   
$\checkmark$&$\checkmark$&$\checkmark$&\textbf{ 79.16}   \\  
\hline
\end{tabular}
\end{center}
\caption{\textbf{Validate the validity of the class local consistency loss and class global consistency loss respectively.} The results of this experiment are obtained by training the ResNet50-based FCN network on the Cityscapes dataset.}
\label{tab:3}
\end{table}

\begin{table}[htbp]
\begin{center}
\begin{tabular}{|l|cc|}
\hline
 $ \alpha$  & $\beta$  & mIou \\
\hline\hline
  \multirow{8}{*}{0.01} & 0.01 & 78.97 \\   
			  & 0.02 & 78.8 \\  
                         & 0.03 & 78.00 \\   
                        & 0.05 & \textbf{79.16} \\  
                        & 0.06 & 78.51 \\   
			& 0.07& 78.03 \\      
                       & 0.1 & 78.07 \\    
			& 0.45 & 78.58 \\  \hline
                0.02 & 0.01 & 78.36 \\
\hline
\end{tabular}
\end{center}
\caption{\textbf{Ablation experiments of class local consistency loss and global consistency loss weights.} The results of this experiment are obtained by training the ResNet50-based FCN network on the Cityscapes dataset with different $\alpha$  and $\beta$ .}
\label{tab:4}
\end{table}

\subsection{Comparison with State-of-the-Art}

\noindent\textbf{Results on ADE20K }   The results with other state-of-the-art methods on ADE20K dataset are summarized in Table \ref{tab:5}. Integration of MDRL into PSPNet\cite{zhao2017pyramid} with ResNet101 as  backbone resulted in 3.41${\%}$ mIoU improvements under multi-scale and flipping test.  Under ResNet-101, our method achieve a mIoU of 46.7${\%}$, which is   0.8${\%}$,0.94${\%}$,1.2${\%}$  mIoU higer than ACNet\cite{fu2019adaptive},CCNet\cite{huang2019ccnet},DMNet\cite{he2019dynamic}. Besides, under the settig of  PSPNet as the semantic aggregation network, MRDL outperforms previous best method MCIBI++ by 0.71${\%}$/0.64${\%}$ mIoU improvements when ResNet-50/ResNet-101 as backbone.  These results show that it is more efficient to use multiple distribution features to describe the intra-class variations .

\begin{table}[htbp]
\begin{center}
\begin{tabular}{|c|ccc|}
\hline

Method & Backbone & Stride & mIoU \\ \hline\hline
                OCNet\cite{yuan2018ocnet} & ResNet-101 & 8× & 45.45\\ 
                CCNet\cite{huang2019ccnet} & ResNet-101 & 8× & 45.76\\ 
                ACNet\cite{fu2019adaptive} & ResNet-101 & 8× & 45.90\\  
                DMNet\cite{he2019dynamic} & ResNet-101 & 8× & 45.50\\ 
                EncNet\cite{zhang2018context} & ResNet-101 & 8× & 44.65\\ 
                PSPNet\cite{zhao2017pyramid} & ResNet-101 & 8× & 43.29\\ 
               PSANet\cite{zhao2018psanet} & ResNet-101 & 8× & 43.77\\
               APCNet\cite{he2019adaptive} & ResNet-101 & 8× & 45.38\\
               OCRNet\cite{yuan2020object} & ResNet-101 & 8× & 45.28\\
               PSPNet\cite{zhao2017pyramid} & ResNet-269 & 8× & 44.94 \\
               UperNet\cite{xiao2018unified} & ResNet-101 & 8× & 44.85\\  
               PSPNet+MCIBI++\cite{jin2022mcibi++}\ddag & ResNet-50 & 8× & 44.72\\
               PSPNet+MCIBI++\cite{jin2022mcibi++}\ddag & ResNet-101 & 8× & 46.06\\ 
               PSPNet+MDRL & ResNet-50 & 8× &\textbf{45.43}\\ 
               PSPNet+MDRL & ResNet-101 & 8× & \textbf{46.70}\\
\hline
\end{tabular}
\end{center}

\caption{\textbf{Segmentation results on ADE20K validation set with other state-of-the-art methods.} Multi-scale and flipping testing are adopted for fair comparison.  \ddag means that this is our re-implemented result under the same experimental  setting.}
\label{tab:5}
\end{table}

\noindent\textbf{Results on Cityscapes }  As shown in Table \ref{tab:6},  we also compared the performance  with other state-of-the-art methods on the validation set of Cityscapes.  The experimental results in the table are all based on single scale testing.  Under ResNet-101 as backbone,   Integration of MDRL into PSPNet\cite{zhao2017pyramid} resulted in 1.25${\%}$ mIoU improvements.  Meanwhile, compare with other state-of-the-art methods, our method outperforms  DNL\cite{yin2020disentangled},OCRNet\cite{yuan2020object},ISNet\cite{jin2021isnet} by 0.61${\%}$,0.31${\%}$,0.45${\%}$ mIoU. Besides,  our method is 0.93${\%}$ higher than MCIBI\cite{jin2021mining} based on ResNet-50 and 0.52${\%}$ higher than MCIBI++\cite{jin2022mcibi++} when ResNet-101 as backbone.

\begin{table}
\begin{center}
\begin{tabular}{|c|ccc|}
\hline
Method & Backbone & Stride & mIoU \\ 
\hline\hline
GCNet\cite{cao2019gcnet} & ResNet-101 & 8× & 79.03\\ 
                PSANet\cite{zhao2018psanet} & ResNet-101 & 8× & 79.31\\ 
                ANN\cite{zhu2019asymmetric} & ResNet-101 & 8× & 77.14\\ 
                NonLocal\cite{wang2018non} & ResNet-101 & 8× & 78.93\\ 
                CCNet\cite{huang2019ccnet} & ResNet-101 & 8× & 78.87\\
                EncNet\cite{zhang2018context} & ResNet-101 & 8× & 78.55\\
                DANet\cite{fu2019dual}   & ResNet-101 & 8× &  80.41 \\
                DNL\cite{yin2020disentangled}  & ResNet-101 & 8× &  80.41 \\
                OCRNet\cite{yuan2020object} & ResNet-101 & 8× &  80.70 \\
                PSPNet\cite{zhao2017pyramid} & ResNet-101 & 8× & 79.76 \\
                ISNet\cite{jin2021isnet} & ResNet-50 & 8× & 79.32 \\
                ISNet\cite{jin2021isnet} & ResNet-101 & 8× & 80.56 \\
                PSPNet+MCIBI\cite{jin2021mining}\ddag& ResNet-50 & 8× & 79.12 \\
                PSPNet+MCIBI++\cite{jin2022mcibi++}\ddag  & ResNet-101 & 8× &  80.49 \\ 
                PSPNet+MDRL  & ResNet-50 & 8× & \textbf{80.05} \\
                PSPNet+MDRL & ResNet-101 & 8× & \textbf{81.01} \\
\hline
\end{tabular}
\end{center}
\caption{\textbf{Segmentation results on Cityscapes validation set with other state-of-the-art methods.} Only single-scale testing is adopted here.  \ddag means that this is our re-implemented result under the same experimental  setting.}
\label{tab:6}
\end{table}

\begin{figure*}[t]
      \centering
	\subfigure[Image]{
		\begin{minipage}[t]{0.2\linewidth}
			\centering
			\includegraphics[width=1.3in,height=1.8cm]{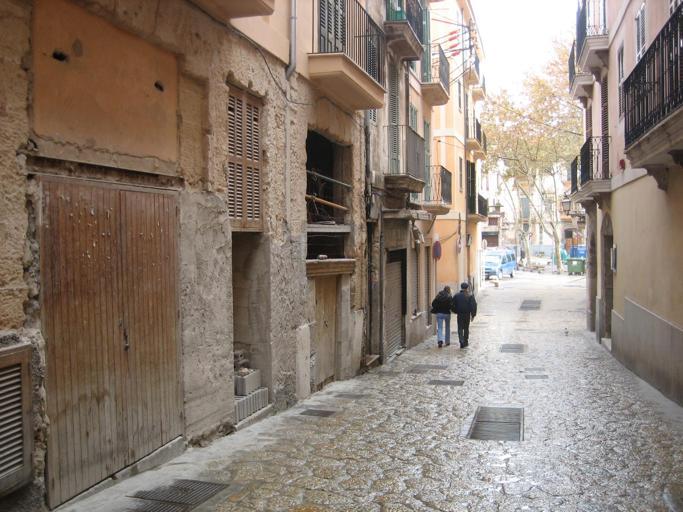}\\
			\vspace{0.02cm}
			\includegraphics[width=1.3in,height=1.8cm]{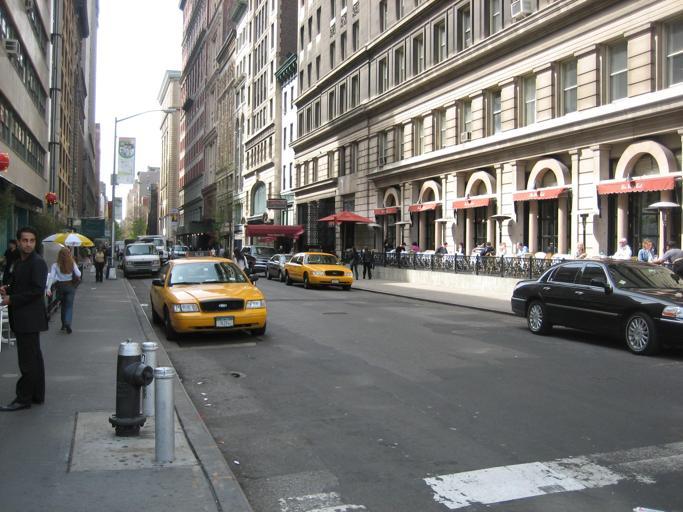}\\
			\vspace{0.02cm}
			\includegraphics[width=1.3in,height=1.8cm]{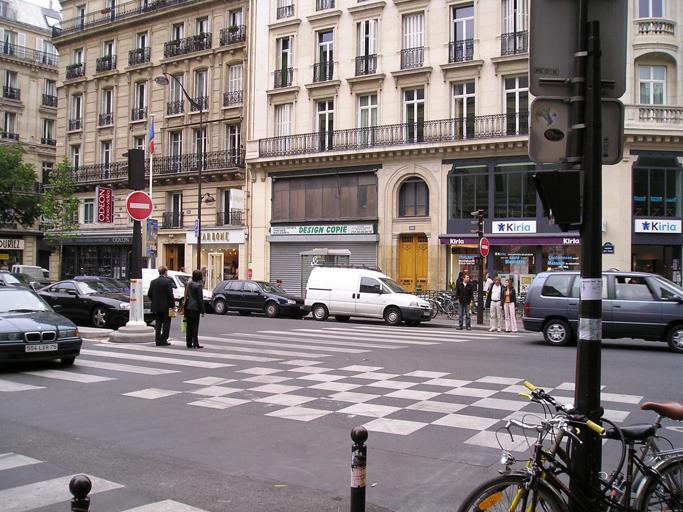}\\
			\vspace{0.02cm}
                       \includegraphics[width=1.3in,height=1.8cm]{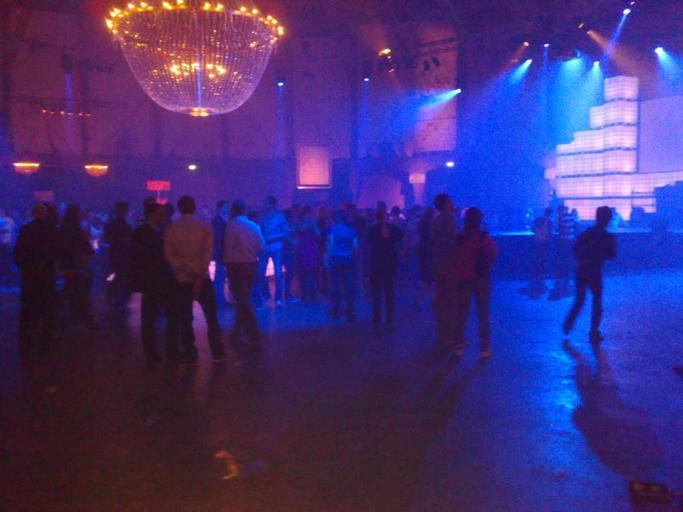}\\
			\vspace{0.02cm}
                       \includegraphics[width=1.3in,height=1.8cm]{ADE20K/image/4_4}\\
			\vspace{0.02cm}
		\end{minipage}%
	}%
\subfigure[GT]{
		\begin{minipage}[t]{0.2\linewidth}
			\centering
			\includegraphics[width=1.3in,height=1.8cm]{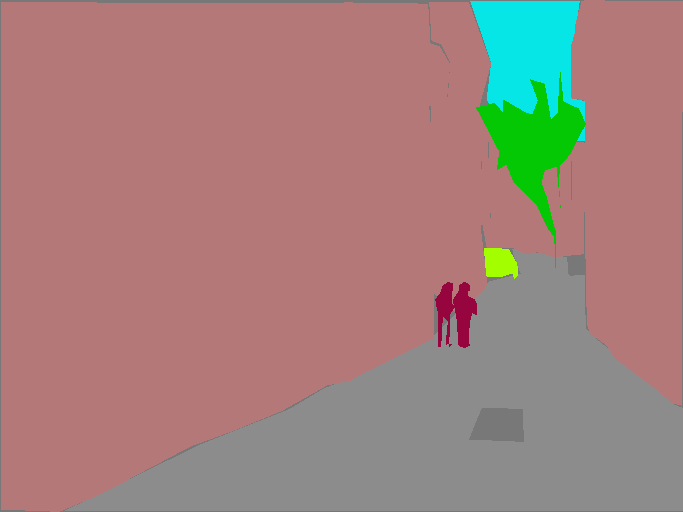}\\
			\vspace{0.02cm}
                   
			\includegraphics[width=1.3in,height=1.8cm]{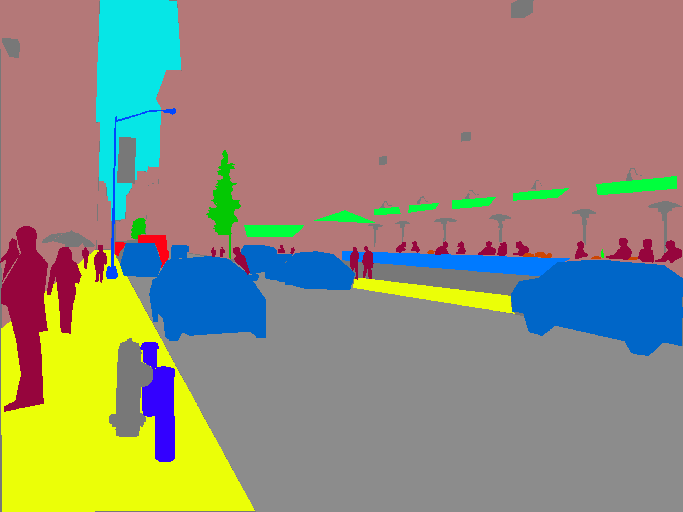}\\
			\vspace{0.02cm}
			\includegraphics[width=1.3in,height=1.8cm]{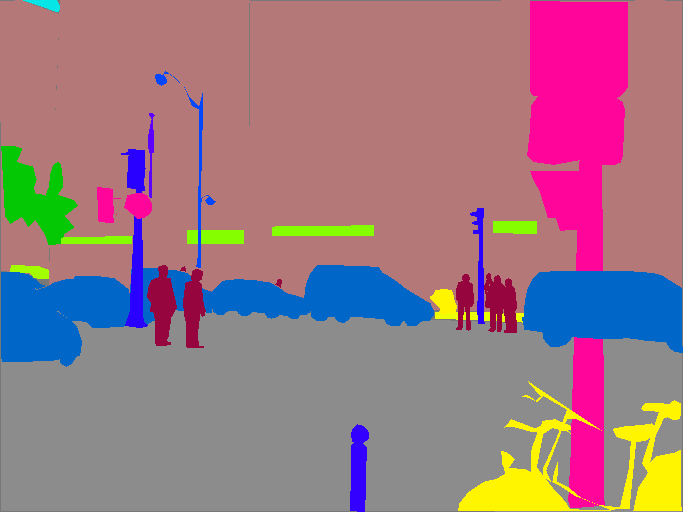}\\
			\vspace{0.02cm}
                       \includegraphics[width=1.3in,height=1.8cm]{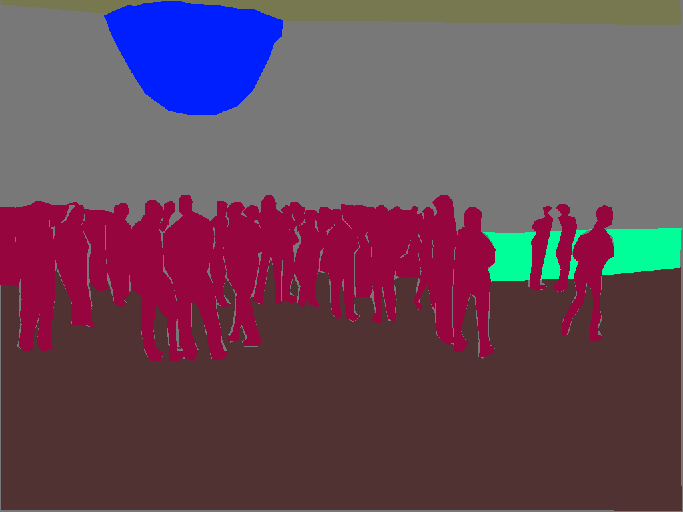}\\
			\vspace{0.02cm}
                       \includegraphics[width=1.3in,height=1.8cm]{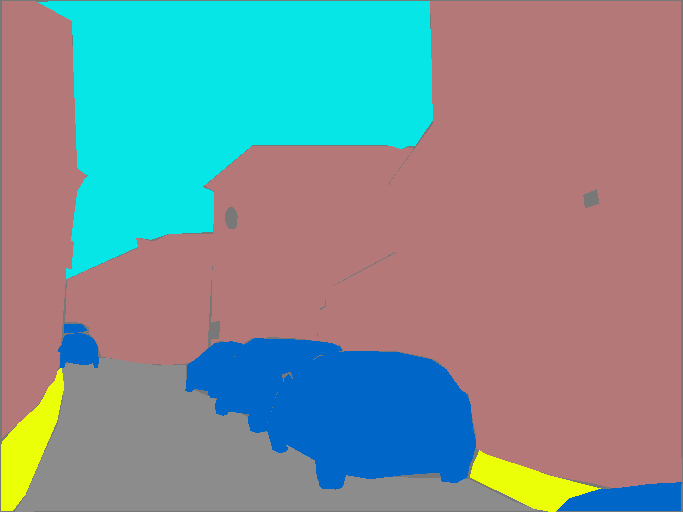}\\
			\vspace{0.02cm}
		\end{minipage}%
	}%
	\subfigure[PSPNet]{
		\begin{minipage}[t]{0.2\linewidth}
			\centering
			\includegraphics[width=1.3in,height=1.8cm]{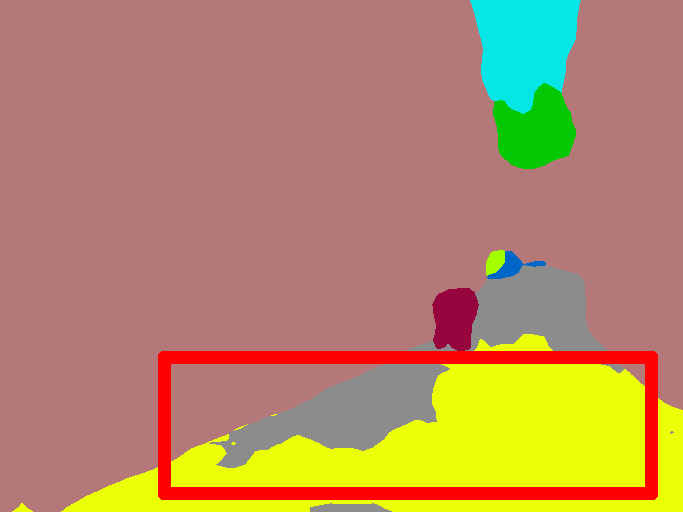}\\
			\vspace{0.02cm}
			\includegraphics[width=1.3in,height=1.8cm]{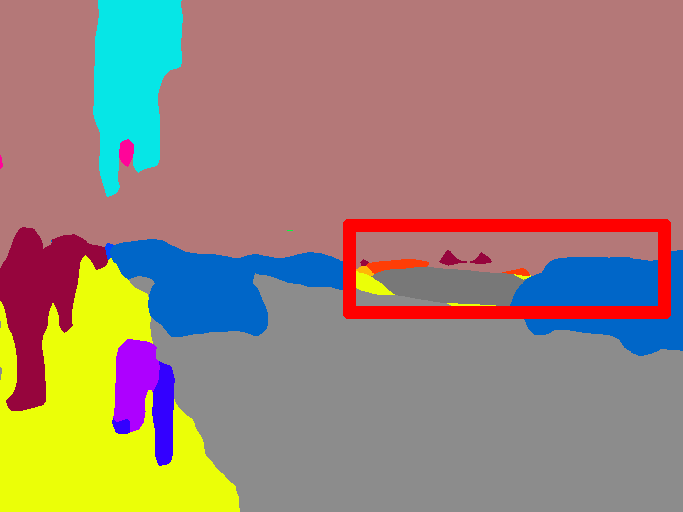}\\
			\vspace{0.02cm}
			\includegraphics[width=1.3in,height=1.8cm]{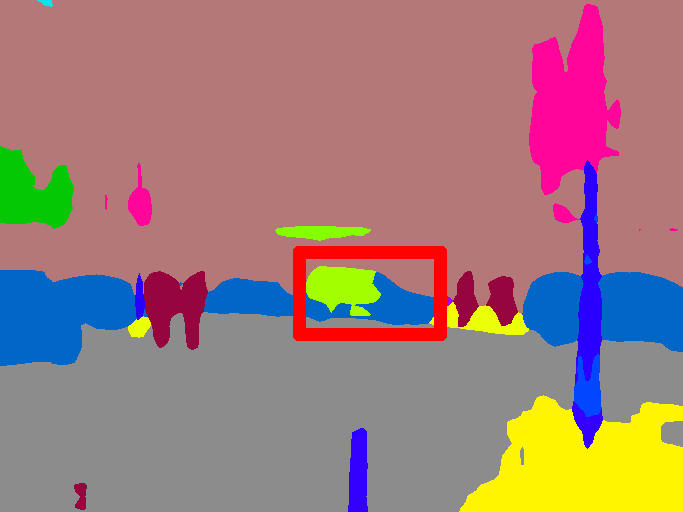}\\
			\vspace{0.02cm}
                       \includegraphics[width=1.3in,height=1.8cm]{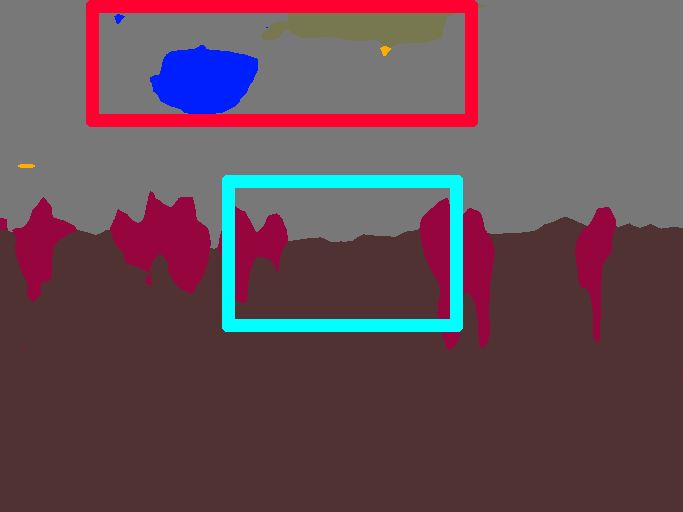}\\
			\vspace{0.02cm}
                       \includegraphics[width=1.3in,height=1.8cm]{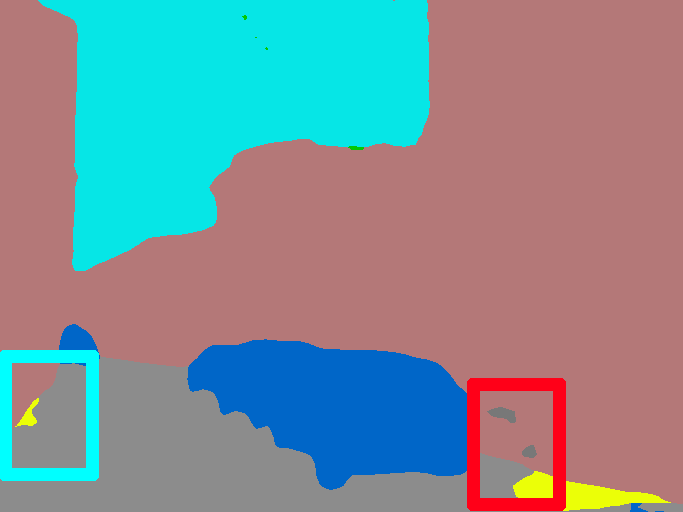}\\
			\vspace{0.02cm}
		\end{minipage}%
	}%
	\subfigure[PSP+MCIBI++]{
		\begin{minipage}[t]{0.2\linewidth}
			\centering
			\includegraphics[width=1.3in,height=1.8cm]{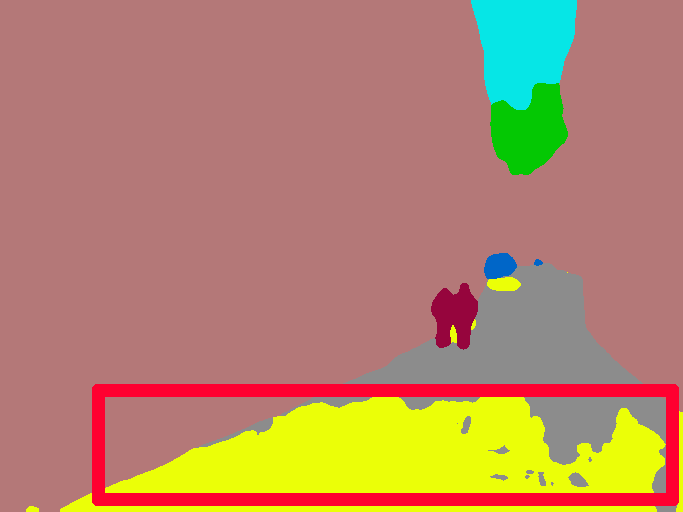}\\
			\vspace{0.02cm}
			\includegraphics[width=1.3in,height=1.8cm]{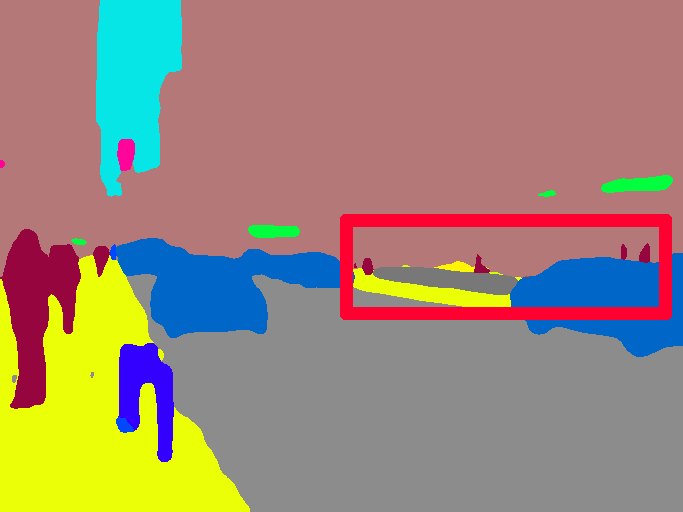}\\
			\vspace{0.02cm}
			\includegraphics[width=1.3in,height=1.8cm]{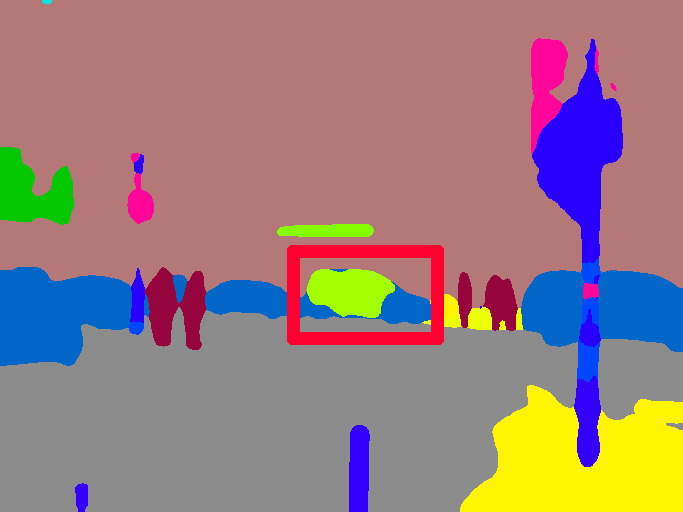}\\
			\vspace{0.02cm}
                       \includegraphics[width=1.3in,height=1.8cm]{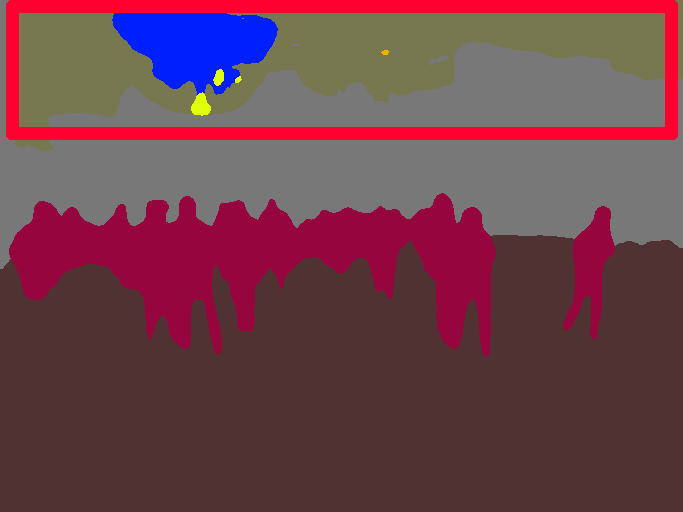}\\
			\vspace{0.02cm}
                       \includegraphics[width=1.3in,height=1.8cm]{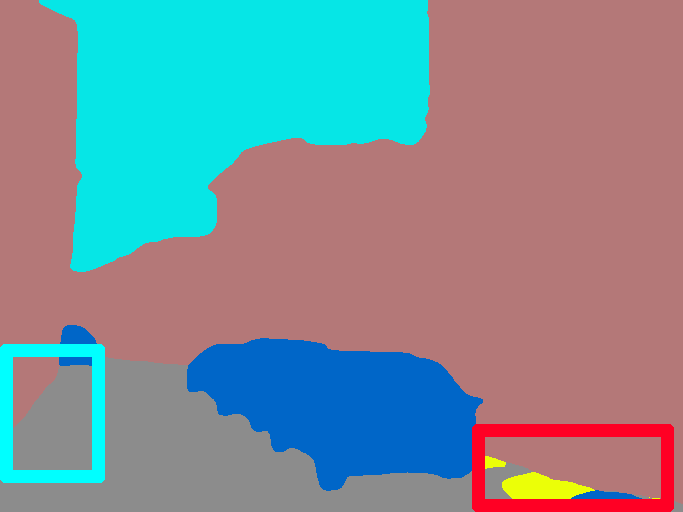}\\
			\vspace{0.02cm}
		\end{minipage}%
	}%
\subfigure[PSP+MDRL]{
		\begin{minipage}[t]{0.2\linewidth}
			\centering
			\includegraphics[width=1.3in,height=1.8cm]{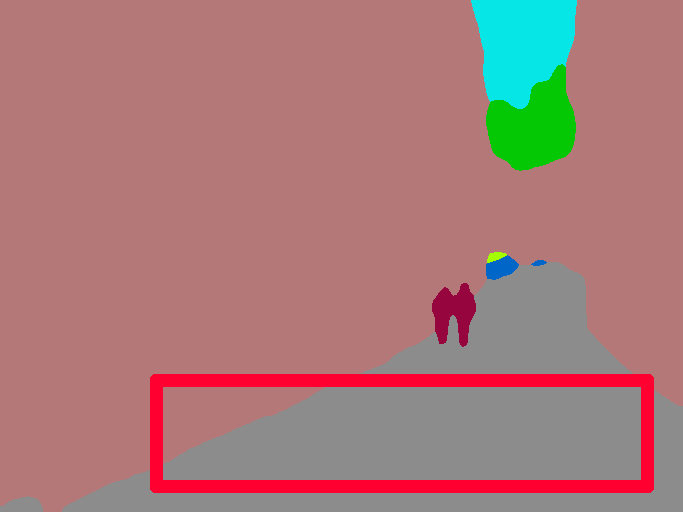}\\
			\vspace{0.02cm}
			\includegraphics[width=1.3in,height=1.8cm]{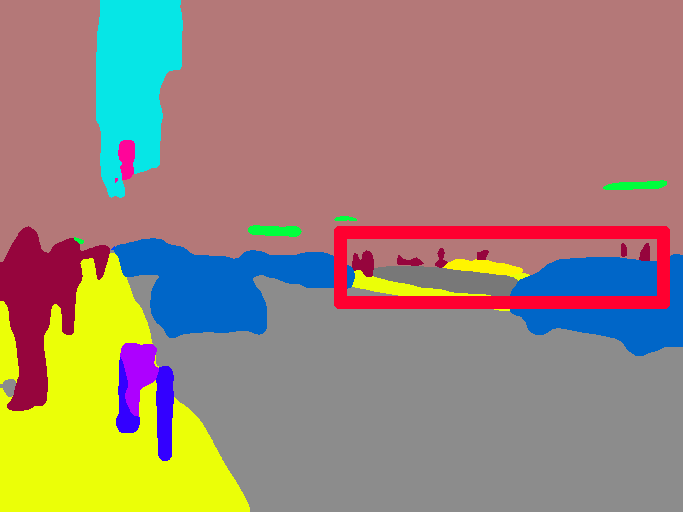}\\
			\vspace{0.02cm}
			\includegraphics[width=1.3in,height=1.8cm]{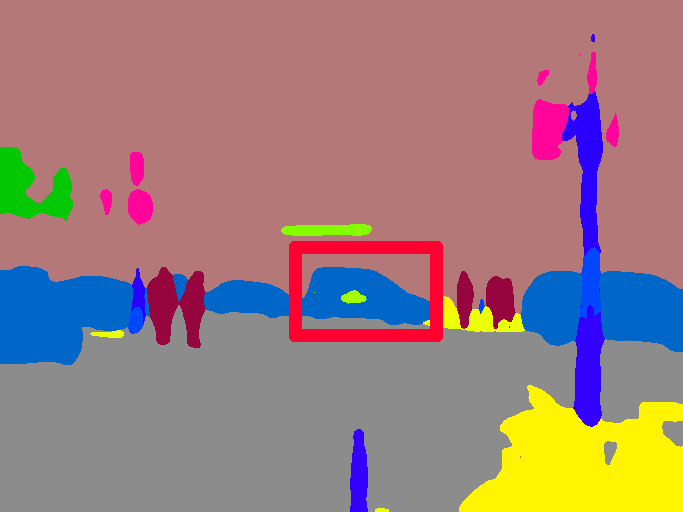}\\
			\vspace{0.02cm}
                       \includegraphics[width=1.3in,height=1.8cm]{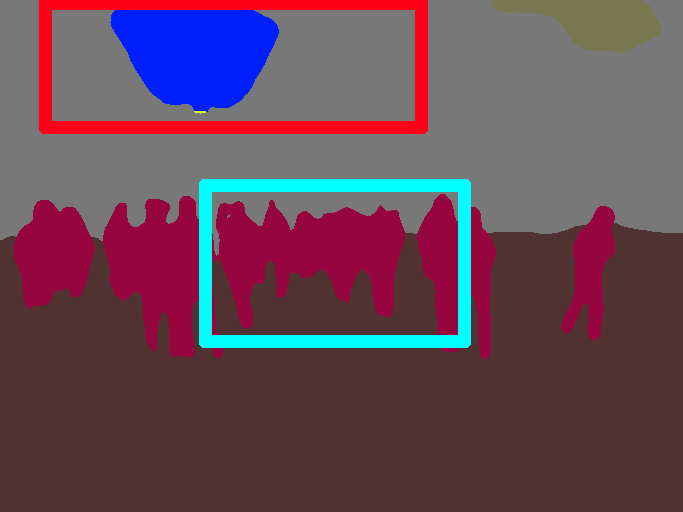}\\
			\vspace{0.02cm}
                       \includegraphics[width=1.3in,height=1.8cm]{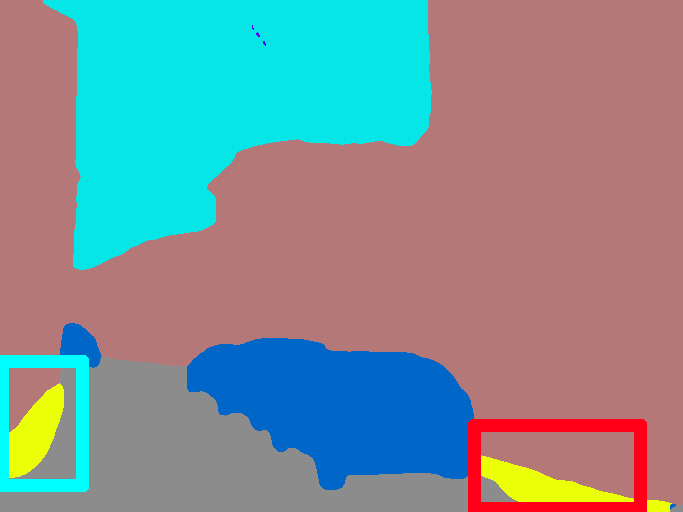}\\
			\vspace{0.02cm}
		\end{minipage}%
	}%
	
	\caption{\textbf{Qualitative results on the validation set of ADE20K.} All models here are trained with ResNet-50 as backbone under the same setting.}
	
	\label{fig:5}

\end{figure*}

\noindent\textbf{Qualitative Results with State-of-the-Arts}  We compared the visualization results with the state-of-the-arts method MCIBI++\cite{jin2022mcibi++} under the same conditions.  As shown in Figure \ref{fig:5} .  Our method performs better between individuals belonging to the same class with large differences. For example, the brightness of the road in the first row is different. Pedestrians located on the right side of the street wearing with different colors and different brightness in the second row.  The third row of cars with different colors.  The fourth line of indoor scenes with different brightness and colors of chandelier and people.
The results show that our method is excellent in segmenting objects belonging to the same class with large differences. Besides,the result in the fifth row shows that our method can better identify two objects that are similar, but belong to different classes.  Qualitative results show the importance of using multiple distributions to describe intra-class variations.

\section{Conclusion}
This paper presents a new perspective to describe class representation  for segmantic segmentation.  We  introduce the multiple distributions to describe  intra-class variations. Then, we propose multiple distributions representation learning to augment the pixel representations. The class multiple distributions consistency strategy is committed to generating class-level multiple distribution features, more fine-grained multiple distribution feature representations are obtained through the feature voting and multi-distribution semantic aggregation module. Qualitative Results demonstrate that our method performs better in classifying objects of the same class but with large differences and objects of different classes but similar. Experimental results show that our approach can be integrated into existing popular segmentation frameworks and achieve  good performance. Meanwhile, our method achieved the state-of-the-art performance on two challenging  datasets.

{\small
\bibliographystyle{plain}
\bibliography{ref.bib}
}

\end{document}